\documentclass[english,aps,manuscript,aip,jcp]{revtex4-1}
\usepackage[T1]{fontenc}
\usepackage[latin9]{inputenc}
\usepackage{babel}
\usepackage{mathtools}
\usepackage{amsmath}
\usepackage{amsthm}
\usepackage{amssymb}
\usepackage{graphicx}
\usepackage[unicode=true]
 {hyperref}

\makeatletter
 
 \@ifundefined{textcolor}{}
 {%
   \definecolor{BLACK}{gray}{0}
   \definecolor{WHITE}{gray}{1}
   \definecolor{RED}{rgb}{1,0,0}
   \definecolor{GREEN}{rgb}{0,1,0}
   \definecolor{BLUE}{rgb}{0,0,1}
   \definecolor{CYAN}{cmyk}{1,0,0,0}
   \definecolor{MAGENTA}{cmyk}{0,1,0,0}
   \definecolor{YELLOW}{cmyk}{0,0,1,0}
 }
\theoremstyle{plain}
\newtheorem{thm}{\protect\theoremname}
  \theoremstyle{plain}
  \newtheorem{assumption}[thm]{\protect\assumptionname}

\usepackage{bbm}
\usepackage{graphicx}

\makeatother

  \providecommand{\assumptionname}{Assumption}
\providecommand{\theoremname}{Theorem}

\begin{document}

\title{Variational Koopman models: slow collective variables and molecular
kinetics from short off-equilibrium simulations}

\author{Hao Wu}
\email{hao.wu@fu-berlin.de}

\selectlanguage{english}%

\affiliation{Department of Mathematics and Computer Science, Freie Universität
Berlin, Arnimallee 6, 14195 Berlin, Germany}

\author{Feliks Nüske}
\email{feliks.nueske@fu-berlin.de}

\selectlanguage{english}%

\thanks{H. Wu and F. Nüske contributed equally to this work.}

\affiliation{Department of Mathematics and Computer Science, Freie Universität
Berlin, Arnimallee 6, 14195 Berlin, Germany}

\author{Fabian Paul}
\email{fab@zedat.fu-berlin.de}

\selectlanguage{english}%

\affiliation{Department of Mathematics and Computer Science, Freie Universität
Berlin, Arnimallee 6, 14195 Berlin, Germany}

\author{Stefan Klus}
\email{stefan.klus@fu-berlin.de}

\selectlanguage{english}%

\affiliation{Department of Mathematics and Computer Science, Freie Universität
Berlin, Arnimallee 6, 14195 Berlin, Germany}

\author{Péter Koltai}
\email{peter.koltai@fu-berlin.de}

\selectlanguage{english}%

\affiliation{Department of Mathematics and Computer Science, Freie Universität
Berlin, Arnimallee 6, 14195 Berlin, Germany}

\author{Frank Noé}
\email{frank.noe@fu-berlin.de}

\selectlanguage{english}%

\affiliation{Department of Mathematics and Computer Science, Freie Universität
Berlin, Arnimallee 6, 14195 Berlin, Germany}
\begin{abstract}
Markov state models (MSMs) and Master equation models are popular
approaches to approximate molecular kinetics, equilibria, metastable
states, and reaction coordinates in terms of a state space discretization
usually obtained by clustering. Recently, a powerful generalization
of MSMs has been introduced, the variational approach (VA) of molecular
kinetics and its special case the time-lagged independent component
analysis (TICA), which allow us to approximate slow collective variables
and molecular kinetics by linear combinations of smooth basis functions
or order parameters. While it is known how to estimate MSMs from trajectories
whose starting points are not sampled from an equilibrium ensemble,
this has not yet been the case for TICA and the VA. Previous estimates
from short trajectories, have been strongly biased and thus not variationally
optimal. Here, we employ Koopman operator theory and ideas from dynamic
mode decomposition (DMD) to extend the VA and TICA to non-equilibrium
data. The main insight is that the VA and TICA provide a coefficient
matrix that we call Koopman model, as it approximates the underlying
dynamical (Koopman) operator in conjunction with the basis set used.
This Koopman model can be used to compute a stationary vector to reweight
the data to equilibrium. From such a Koopman-reweighted sample, equilibrium
expectation values and variationally optimal reversible Koopman models
can be constructed even with short simulations. The Koopman model
can be used to propagate densities, and its eigenvalue decomposition
provide estimates of relaxation timescales and slow collective variables
for dimension reduction. Koopman models are generalizations of Markov
state models, TICA and the linear VA and allow molecular kinetics
to be described without a cluster discretization.
\end{abstract}
\maketitle

\section{Introduction}

With the ability to generate extensive and high-throughput molecular
dynamics (MD) simulations \cite{ShirtsPande_Science2000_FoldingAtHome,Phllips_JCC05_NAMD,HarveyDeFabritiis_JCTC09_ACEMD,BuchEtAl_JCIM10_GPUgrid,Shaw_Science10_Anton,EastmanEtAl_JCTC13_OpenMM,GrandWalker_CPC13_MDGPU,PronkEtAl_Bioinf13_Gromacs4.5,DoerrEtAl_JCTC16_HTMD},
the spontaneous sampling of rare-events such as protein folding, conformational
changes and protein-ligand association have become accessible \cite{NoeSchuetteReichWeikl_PNAS09_TPT,BuchFabritiis_PNAS11_Binding,BowmanVoelzPande_JACS11_FiveHelixBundle-TripletQuenching,SadiqNoeFabritiis_PNAS12_HIV,SilvaEtAl_PNAS14_RNAPolymeraseII,ShuklaPande_NatCommun14_SrcKinase,PlattnerNoe_NatComm15_TrypsinPlasticity,ReuboldEtAl_Nature15_DynaminTetramer}.
Markov state models (MSMs) \cite{SchuetteFischerHuisingaDeuflhard_JCompPhys151_146,SwopePiteraSuits_JPCB108_6571,NoeHorenkeSchutteSmith_JCP07_Metastability,ChoderaEtAl_JCP07,Bowman_JCP09_Villin,PrinzEtAl_JCP10_MSM1,SchuetteEtAl_JCP11_Milestoning,BowmanPandeNoe_MSMBook},
Master-equation models \cite{ChekmarevIshidaLevi_JPCB04_MasterEq,SriramanKevrekidisHummer_JPCB109_6479,BucheteHummer_JPCB08}
and closely related approaches \cite{NoeEtAl_PMMHMM_JCP13,WuNoe_MMS14_TRAM1,RostaHummer_DHAM,WuNoe_JCP15_GMTM,WuEtAL_PNAS16_TRAM}
have emerged as powerful frameworks for the analysis of extensive
MD simulation data. These methods do not require a very specific \emph{a
priori} definition of relevant reaction coordinates \cite{SarichNoeSchuette_MMS09_MSMerror,PrinzEtAl_JCP10_MSM1}.
Furthermore, they allow a large variety of mechanistic information
to be extracted \cite{NoeSchuetteReichWeikl_PNAS09_TPT,MetznerSchuetteVandenEijnden_TPT,BerezhkovskiiHummerSzabo_JCP09_Flux},
and experimental observables to be computed and structurally interpreted
\cite{BucheteHummer_JPCB08,NoeEtAl_PNAS11_Fingerprints,BowmanVoelzPande_JACS11_FiveHelixBundle-TripletQuenching,KellerPrinzNoe_ChemPhysReview11,Zhuang_JPCB11_MSM-IR,LindnerEtAl_JCP13_NeutronScatteringI}.
Finally, they provide a direct approximation of the dynamic modes
describing the slow conformational changes that are identical or closely
related to the so-called reaction coordinates, depending on which
notion of that term is employed \cite{PetersTrout_JCP06_ReactionCoordinateOptimization,Peters_jcp06_committorerrorestimation,ChoderaPande_PRL11_SplittingProbabilities,RohrdanzClementi_JCP134_DiffMaps,RohrdanzEtAl_AnnRevPhysChem13_MountainPasses,McGibbonPande_Arxiv16_SparseTICA}.
An especially powerful feature of MSMs and similar approaches is that
the transition probabilities $p_{ij}(\tau)$, i.e. the probability
that the trajectory is found in a set $A_{j}$ a time lag $\tau$
after it has been found in a set $A_{i}$, 
\[
p_{ij}(\tau)=\mathrm{Prob}\left[\mathbf{x}_{t+\tau}\in A_{j}\mid\mathbf{x}_{t}\in A_{i}\right],
\]
is a conditional transition probability. $p_{ij}(\tau)$ can be estimated
without bias even if the trajectory is not initiated from a global,
but only a local equilibrium distribution \cite{PrinzEtAl_JCP10_MSM1}.
Consequently, given $c_{ij}(\tau)$ transition events between states
$i$ and $j$ at lag time $\tau$, the maximum likelihood estimator
of the transition probability can be easily shown to be
\begin{equation}
p_{ij}(\tau)=\frac{c_{ij}(\tau)}{\sum_{k}c_{ik}(\tau)},\label{eq:P_nonrev}
\end{equation}
i.e. the fraction of the number of transitions to $j$ conditioned
on starting in $i$. This conditionality is a key reason why MSMs
have become popular to analyze short distributed simulations that
are started from arbitrary configurations whose relationship to the
equilibrium distribution is initially unknown.

However, when estimating (\ref{eq:P_nonrev}) from simulation data,
one does not generally obtain a time-reversible estimate, i.e. the
stationary probabilities of the transition matrix, $\pi_{i}$, will
usually not fulfill the detailed balance equations $\pi_{i}p_{ij}=\pi_{j}p_{ji}$,
even if the underlying dynamics are microscopically time-reversible.
Compared to a reversible transition matrix, a transition matrix with
independent estimates of $p_{ij}$ and $p_{ji}$ has more free parameters,
resulting in larger statistical uncertainties, and may possess complex-valued
eigenvalues and eigenvectors, which limits the application of some
analysis tools designed for equilibrium molecular processes \cite{TrendelkampSchroerEtAl_InPrep_revMSM}.
Since most molecular dynamics simulations are in thermal equilibrium
and thus fulfill at least a generalized microscopic reversibility
(Appendix B in \cite{BittracherKoltaiJunge_SIAM15_Pseudogenerators}),
it is desirable to force $p_{ij}$ to fulfill detailed balance, which
both reduces statistical uncertainty and enforces a real-valued spectrum
\cite{Noe_JCP08_TSampling,TrendelkampSchroerEtAl_InPrep_revMSM}.
In old studies, the pragmatic solution to this problem was often to
symmetrize the count matrix, i.e. to simply set $c_{ij}^{\mathrm{sym}}=c_{ij}+c_{ji}$,
which is equivalent to evaluating the simulation trajectory forward
and backward, and which leads to a transition matrix with detailed
balance when inserted into (\ref{eq:P_nonrev}). However, it has been
known since at least 2008 that this estimator is strongly biased,
and therefore reversible maximum likelihood and Bayesian estimators
have been developed \cite{Noe_JCP08_TSampling,BucheteHummer_JPCB08,Bowman_JCP09_Villin,PrinzEtAl_JCP10_MSM1,TrendelkampNoe_JCP13_EfficientSampler,TrendelkampSchroerEtAl_InPrep_revMSM}.
These algorithms formulate the estimation problem as an optimization
or sampling problem of the transition matrix constrained to fulfill
detailed balance. The idea of these algorithms becomes clear when
writing the reversible maximum likelihood estimator in two subsequent
steps, as demonstrated in \cite{TrendelkampSchroerEtAl_InPrep_revMSM}:
\begin{enumerate}
\item \emph{Reweighting}: Estimate the stationary distribution $\pi_{i}$
given all transition counts $c_{ij}$ under the detailed balance condition. 
\item \emph{Estimation}: Insert $\pi_{i}$ and $c_{ij}$ into an equation
for the reversible transition matrix to obtain a maximum likelihood
estimate of $p_{ij}$.
\end{enumerate}
Recently, a powerful extension to the Markov modeling framework has
been introduced: the variational approach (VA) to approximate the
slow components of reversible Markov processes \cite{NoeNueske_MMS13_VariationalApproach}.
Due to its relevance for molecular dynamics, it has also been referred
to as VA for molecular kinetics \cite{PerezEtAl_JCP13_TICA,NueskeEtAl_JCTC14_Variational}
or VA for conformation dynamics \cite{PerezEtAl_JCP13_TICA,VitaliniNoeKeller_JCTC15_BasisSet}.
It has been known for many years that Markov state models are good
approximations to molecular kinetics if their largest eigenvalues
and eigenvectors approximate the eigenvalues and eigenfunctions of
the Markov operator governing the full-phase space dynamics \cite{Huisinga_AnnApplProbab04_PhaseTransitions,SchuetteFischerHuisingaDeuflhard_JCompPhys151_146,SarichNoeSchuette_MMS09_MSMerror},
moreover the first few eigenvalues and eigenvectors are sufficient
to compute almost all stationary and kinetic quantities of interest
\cite{DeuflhardWeber_PCCA,KubeWeber_JCP07_CoarseGraining,NoeEtAl_PNAS11_Fingerprints,KellerPrinzNoe_ChemPhysReview11,CameronVandenEijnden_JSP14_FlowsInComplexNetworks}.
The VA has generalized this idea beyond discrete states and formulated
the approximation problem of molecular kinetics in terms of an approach
that is similar to the variational approach in quantum mechanics \cite{NoeNueske_MMS13_VariationalApproach,PerezEtAl_JCP13_TICA,NueskeEtAl_JCTC14_Variational}.
It is based on the following variational principle: If we are given
a set of $n$ orthogonal functions of state space, and evaluate the
autocorrelations of the molecular dynamics in these functions at lag
time $\tau$, these will give us lower bounds to the true eigenvalues
$\lambda_{1}(\tau),\,...,\,\lambda_{n}(\tau)$ of the Markov operator,
equivalent to an underestimate of relaxation timescales and an overestimate
of relaxation rates. Only if the $n$ functions used are the eigenfunctions
themselves, then their autocorrelations will be maximal and identical
to the true eigenvalues $\lambda_{1}(\tau),\,...,\,\lambda_{n}(\tau)$.
Note that this statement is true in the correct statistical limit
- for finite data, the variational bound can be violated by problems
in the estimation procedure. Sources of violation include systematic
estimator bias, which is addressed in this work, and overfitting,
which can be addressed by cross-validation \cite{McGibbonPande_JCP15_CrossValidation}.

This principle allows to formulate variational optimization algorithms
to approximate the eigenvalues and eigenfunctions of the Markov operator.
The linear VA proceeds as follows:
\begin{enumerate}
\item Fix an arbitrary basis set $\boldsymbol{\chi}=\left[\chi_{1}(\mathbf{x}),\,...,\,\chi_{n}(\mathbf{x})\right]$
and evaluate the values of all basis functions for all sampled MD
configurations $\mathbf{x}$.
\item Estimate two covariance matrices, the covariance matrix $\mathbf{C}(0)$,
and the time-lagged covariance matrix $\mathbf{C}(\tau)$ from the
basis-set-transformed data.
\item Solve a generalized eigenvalue problem involving both $\mathbf{C}(0)$
and $\mathbf{C}(\tau)$, and obtain estimates for the eigenvalues
and the optimal representation of eigenfunctions as a linear combination
of basis functions.
\end{enumerate}
Note that the functions $\boldsymbol{\chi}$ can be arbitrary nonlinear
functions in the original coordinates $\mathbf{x}$, which allows
complex nonlinear dynamics to be encoded even within this linear optimization
framework. The variational approach has spawned a variety of follow-up
works, for example it has been shown that the algorithm called blind
source separation, time-lagged or time-structure based independent
component analysis (TICA) established in signal processing and machine
learning \cite{Molgedey_94,ZieheMueller_ICANN98_TDSEP,HyvaerinenKarhunenOja_ICA_Book}
is a special case of the VA \cite{PerezEtAl_JCP13_TICA}. TICA is
now widely used in order to reduce the dimensionality of MD data sets
to a few slow collective coordinates, in which MSMs and other kinetic
models can be built efficiently \cite{PerezEtAl_JCP13_TICA,SchwantesPande_JCTC13_TICA,NaritomiFuchigami_JCP11_TICA}.
The VA has been used to generate and improve guesses of collective
reaction coordinates \cite{BoninsegnaEtAl_JCTC15_VariationalDM,McGibbonPande_Arxiv16_SparseTICA}.
A VA-based metric has been defined which transforms the simulation
data into a space in which Euclidean distance corresponds to kinetic
distance \cite{NoeClementi_JCTC15_KineticMap,NoeClementi_JCTC16_KineticMap2}.
The importance of meaningful basis sets has been discussed, and a
basis for peptide dynamics has been proposed in \cite{VitaliniNoeKeller_JCTC15_BasisSet}.
Kernel versions of TICA have been proposed \cite{Harmeling_NeurComp03_kernelTDSEP,SchwantesPande_JCTC15_kTICA}
and nonlinear deep versions have been proposed based on tensor approximations
of products of simple basis functions \cite{NueskeEtAl_JCP15_Tensor}.
Finally, the variational principle ranks kinetic models by the magnitude
of their largest eigenvalues or derived quantities \cite{NoeNueske_MMS13_VariationalApproach},
which can be used to select hyper-parameters such as the basis functions
$\boldsymbol{\chi}$, or the number of states in a Markov state model
\cite{McGibbonPande_JCP15_CrossValidation,SchererEtAl_JCTC15_EMMA2}.

Despite the popularity of the VA and TICA, their estimation from MD
data is still in the stage that MSMs had been about a decade ago:
A direct estimation of covariance matrices will generally provide
a non-symmetric $\mathbf{C}(\tau)$ matrix and complex eigenvalues/eigenfunction
estimates that are not consistent with reversible molecular dynamics.
In order to avoid this problem, the current state of the art is to
enforce the symmetrization of covariance matrices directly \cite{PerezEtAl_JCP13_TICA,SchwantesPande_JCTC13_TICA,SchwantesPande_JCTC15_kTICA}.
In lack of a better estimator, this approach is currently used also
with short MD simulations from distributed computing despite the fact
that the resulting timescales and eigenfunctions may be biased and
misleading. This problem is addressed in the present paper. 

The algorithm of the linear VA \cite{NoeNueske_MMS13_VariationalApproach,NueskeEtAl_JCTC14_Variational}
is identical to more recently proposed extended dynamic mode decomposition
(EDMD) \cite{williams2015data}, which is based on dynamic mode decomposition
(DMD) \cite{SchmidSesterhenn_APS08_DMD,RowleyEtAl_JFM09_DMDSpectral,Schmid_JFM10_DMD,TuEtAl_JCD14_ExactDMD}.
However, while the VA has been derived for reversible and stationary
dynamics, EDMD has been developed in the context of dynamical systems
and fluid mechanics, where data is often nonreversible and non-stationary.
Mathematically, these methods are based on the eigenvalue decomposition
of the Koopman operator, which provides a theoretical description
of non-stationary and non-equilibrium dynamics \cite{mezic2005spectral,KlusKoltaiSchuette_ApproximationKoopman}.
In the present paper, this theory is used in order to formulate robust
equilibrium estimators for covariance matrices, even if the simulation
data are generated in many short simulations that are not distributed
according to equilibrium. Based on these estimates, a Koopman model
is computed - a matrix model that approximates the dynamics of the
Koopman operator on the basis functions used. Koopman models are proper
generalizations of Markov state models - they do not rely on a state
space clustering, but can still be used to propagate densities in
time, and their eigenvalues and eigenvectors provide estimates of
the equilibrium relaxation timescales and slow collective variables.
We propose a reversible Koopman model estimator that proceeds analogously
to reversible MSM estimation:
\begin{enumerate}
\item \emph{Reweighting}: Estimate a reweighting vector $u_{i}$ with an
entry for each basis function given the covariance matrices $\hat{\mathbf{C}}(0)$
and $\hat{\mathbf{C}}(\tau)$ that have been empirically estimated
without symmetrization.
\item \emph{Estimation}: Insert $u_{i}$ and $\hat{\mathbf{C}}(0),\:\hat{\mathbf{C}}(\tau)$
into an equation for the symmetric equilibrium estimates of $\mathbf{C}(0)$
and $\mathbf{C}(\tau)$. Then compute a Koopman model, and from its
eigenvalue decomposition the relaxation timescales and slow collective
variables.
\end{enumerate}
In addition to this result, the reweighting vector $u_{i}$ allows
us to approximate \emph{any} equilibrium estimate in terms of a linear
combination of our basis functions from off-equilibrium data. The
estimator is asymptotically unbiased in the limits of many short trajectories
and an accurate basis set.

The new methods are illustrated on toy examples with stochastic dynamics
and a benchmark protein-ligand binding problem. The methods described
in this article are implemented in PyEMMA version 2.3 or later (\href{http://www.pyemma.org}{www.pyemma.org})
\cite{SchererEtAl_JCTC15_EMMA2}.

\section{Variational Approach of molecular kinetics\label{sec:Variational-Approach}}

The VA is an algorithmic framework to approximate the slow components
of molecular dynamics - also called conformation dynamics or molecular
kinetics - from data. It consists of two main ingredients: (1) a variational
principle that provides a computable score of a model of the slow
components, and (2) an algorithm based on the variational principle
that estimates slow components from simulation data.

\subsection{Variational principle of conformation dynamics}

Simulations of molecular dynamics (MD) can be modeled as realizations
of an ergodic and time-reversible Markov process $\{\mathbf{x}_{t}\}$
in a phase space $\Omega$. $\mathbf{x}_{t}$ contains all variables
that determine the conformational progression after time $t$ (e.g.,
positions and velocities of all atoms). The time evolution of the
probability distribution $p_{t}\left(\mathbf{x}\right)$ of the molecular
ensemble can be decomposed into a set of relaxation processes as
\begin{equation}
p_{t+\tau}\left(\mathbf{x}\right)=\sum_{i=1}^{\infty}\mathrm{e}^{-\frac{\tau}{t_{i}}}\,\mu\left(\mathbf{x}\right)\psi_{i}\left(\mathbf{x}\right)\,\left\langle \psi_{i},p_{t}\right\rangle ,\label{eq:p-decomposition}
\end{equation}
where $\mu\left(\mathbf{x}\right)$ is the stationary (Boltzmann)
density of the system, $t_{i}$ are relaxation timescales sorted in
decreasing order, $\psi_{i}$ are eigenfunctions of the backward operator
or Koopman operator of $\{\mathbf{x}_{t}\}$ with eigenvalues $\lambda_{i}\left(\tau\right)=\mathrm{e}^{-\frac{\tau}{t_{i}}}$
(see Section \ref{subsec:Koopman-Ooperator-description}), and the
inner product is defined as $\left\langle \psi_{i},p_{t}\right\rangle =\int\mathrm{d}\mathbf{x}\ \psi_{i}\left(\mathbf{x}\right)p_{t}\left(\mathbf{x}\right)$.
The first spectral component is given by the constant eigenfunction
$\psi_{1}\left(\mathbf{x}\right)=\mathbbm1\left(\mathbf{x}\right)\equiv1$
and infinite timescale $t_{1}=\infty>t_{2}$ corresponding to the
stationary state of the system. According to this decomposition, the
$m$ dominant eigenfunctions $\psi_{1},\ldots,\psi_{m}$ can be interpreted
as $m$ slow collective variables, which characterize the behavior
of a molecular system on large time scales $\tau\gg t_{m+1}$.

The eigenvalues and eigenfunctions can also be formulated by the following
variational principle \cite{NoeNueske_MMS13_VariationalApproach,NueskeEtAl_JCTC14_Variational}:
\emph{For any $m\ge1$, the first $m$ eigenfunctions $\psi_{1},\ldots,\psi_{m}$
are the solution of the following optimization problem
\begin{align}
R_{m}=\max_{f_{1},\ldots,f_{m}} & \sum_{i=1}^{m}\mathbb{E}_{\mu}\left[f_{i}\left(\mathbf{x}_{t}\right)f_{i}\left(\mathbf{x}_{t+\tau}\right)\right],\label{eq:variational-principle}\\
\mathrm{s.t.}\quad & \mathbb{E}_{\mu}\left[f_{i}\left(\mathbf{x}_{t}\right)^{2}\right]=1,\nonumber \\
 & \mathbb{E}_{\mu}\left[f_{i}\left(\mathbf{x}_{t}\right)f_{j}\left(\mathbf{x}_{t}\right)\right]=0,\text{ for }i\neq j,\nonumber 
\end{align}
where $\mathbb{E}_{\mu}\left[\cdot\right]$ denotes the expected value
with $\mathbf{x}_{t}$ sampled from the stationary density $\mu$
and the maximum value is the generalized Rayleigh quotient, or Rayleigh
trace $R_{m}=\sum_{i=1}^{m}\lambda_{i}$. }Therefore, for every other
set of functions that aims at approximating the true eigenfunctions,
the eigenvalues will be underestimated, and we can use this variational
principle in order to search for the best approximation of eigenfunctions
and eigenvalues.

\subsection{Linear variational approach\label{subsec:Linear-variational-approach}}

In this paper, we focus on algorithms that approximate the eigenfunctions
in the spectral decomposition (\ref{eq:p-decomposition}) by a linear
combination of real-valued basis functions, also called feature functions,
$\boldsymbol{\chi}=\left(\chi_{1},\ldots,\chi_{m}\right)^{\top}$.
Thus, we make the Ansatz: 

\begin{equation}
f_{i}(\mathbf{x})=\sum_{j=1}^{m}b_{ij}\chi_{j}(\mathbf{x})=\mathbf{b}_{i}^{\top}\boldsymbol{\chi}(\mathbf{x})\label{eq:eigenfunction-approximation}
\end{equation}
with expansion coefficients $\mathbf{b}_{i}=\left(b_{i1},\ldots,b_{im}\right)^{\top}$.
Note that the functions $\boldsymbol{\chi}$ are generally nonlinear
in $\mathbf{x}$, however we will call the resulting algorithm a linear
VA because it is linear in the variables $\mathbf{b}_{i}$.

\textbf{Linear VA algorithm}: By solving (\ref{eq:variational-principle})
with the Ansatz (\ref{eq:eigenfunction-approximation}), we obtain
the linear VA to optimally approximate eigenvalues $\lambda_{i}$
and eigenfunctions $\psi_{i}$ \cite{NoeNueske_MMS13_VariationalApproach,NueskeEtAl_JCTC14_Variational}.
We first estimate the equilibrium covariance and time-lagged covariance
matrices of the basis functions:
\begin{eqnarray}
\mathbf{C}\left(0\right) & = & \mathbb{E}_{\mu}\left[\boldsymbol{\chi}\left(\mathbf{x}_{t}\right)\boldsymbol{\chi}\left(\mathbf{x}_{t}\right)^{\top}\right],\label{eq:C0_eq_expectation}\\
\mathbf{C}\left(\tau\right) & = & \mathbb{E}_{\mu}\left[\boldsymbol{\chi}\left(\mathbf{x}_{t}\right)\boldsymbol{\chi}\left(\mathbf{x}_{t+\tau}\right)^{\top}\right],\label{eq:Ct_eq_expectation}
\end{eqnarray}
then the solution of the generalized eigenvalue problem 
\begin{equation}
\mathbf{C}\left(\tau\right)\mathbf{B}=\mathbf{C}\left(0\right)\mathbf{B}\hat{\boldsymbol{\Lambda}},\label{eq:variational-eigenproblem}
\end{equation}
provides the optimal approximation to eigenvalues $\hat{\boldsymbol{\Lambda}}=\mathrm{diag}\left(\hat{\lambda}_{1},\ldots,\hat{\lambda}_{m}\right)$
and expansion coefficient $\mathbf{B}=\left(\mathbf{b}_{1},\ldots,\mathbf{b}_{m}\right)$.
Inserting these coefficients into (\ref{eq:eigenfunction-approximation})
results in the approximated eigenfunctions \cite{NoeNueske_MMS13_VariationalApproach,NueskeEtAl_JCTC14_Variational}.
An important observation is that ((\ref{eq:variational-eigenproblem}))
is formally equivalent to the eigenvalue decomposition of $\mathbf{K}=\mathbf{C}\left(0\right)^{-1}\mathbf{C}\left(\tau\right)$
if $\mathbf{C}\left(0\right)$ has full rank. $\mathbf{K}$ is the
Koopman model that is the central object of the present paper and
will provide the basis for constructing equilibrium estimates from
short simulations.

\begin{figure}
\begin{centering}
\includegraphics[width=0.7\textwidth]{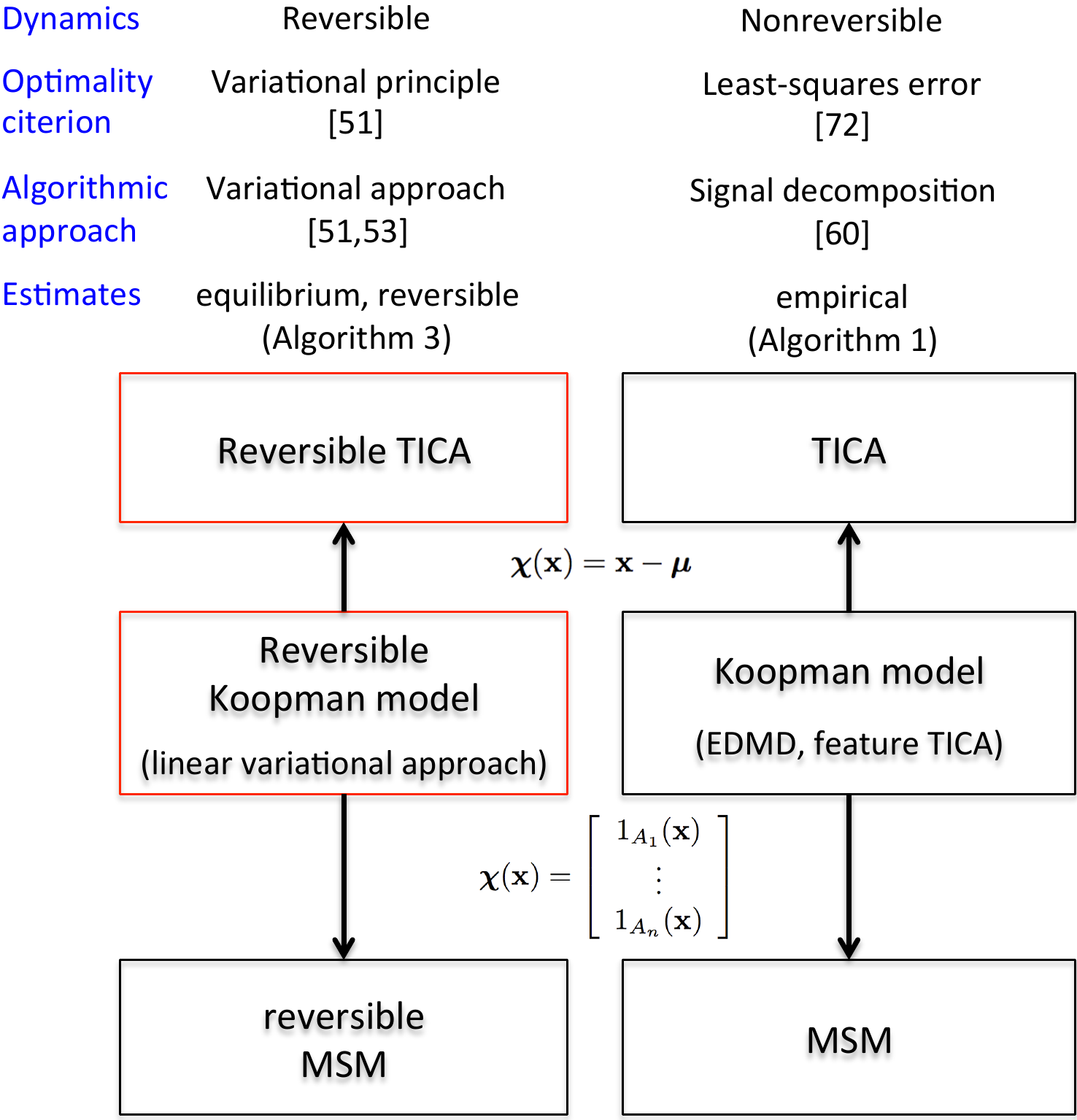}
\par\end{centering}
\caption{\label{fig:relationships}Relationships between different methods
for estimating the slow components of molecular kinetics: Methods
for reversible dynamics are based on the variational principle, leading
to the variational approach \cite{NoeNueske_MMS13_VariationalApproach,NueskeEtAl_JCTC14_Variational}.
Methods for nonreversible dynamics can be derived by minimizing the
least-squares error between the predicted and the observed dynamics,
and lead to a signal decomposition algorithm also called blind source
separation \cite{Molgedey_94}. Interestingly, the nonreversible estimates
can also be obtained by implementing the using empirical estimates
instead of reversible equilibrium estimates of covariance matrices.
Amongst the nonreversible methods, the most general is the Koopman
model estimation (Algorithm 1 here), as they employ linear combinations
of arbitrary basis functions. Their eigenvalue decompositions are
known as EDMD and TICA in feature space. Regular TICA can be derived
if the basis functions are linear in the original coordinates, and
MSMs are obtained by using characteristic functions as basis. Amongst
reversible methods, the variational approach leading to a reversible
Koopman model (Algorithm 3 here) is the most general, and reversible
TICA / reversible MSM estimation methods can be derived by appropriate
basis set choices. The methods in red boxes are derived in this paper,
and the key to these algorithms is the ability to conduct a reversible
equilibrium estimate of covariance matrices for general basis sets
(Algorithm 2 here).}
\end{figure}

The linear VA algorithm provides a general framework for the finite-dimensional
approximation of spectral components of conformation dynamics, and
two widely used analysis methods, time-lagged independent component
analysis (TICA) \cite{Molgedey_94,PerezEtAl_JCP13_TICA,SchwantesPande_JCTC13_TICA}
and Markov state models (MSMs) \cite{PrinzEtAl_JCP10_MSM1}, are both
special cases of the linear VA, see also Fig. \ref{fig:relationships}.

\noindent\textbf{TICA: }In TICA, basis functions are mean-free molecular
coordinates, $\boldsymbol{\chi}(\mathbf{x})=\mathbf{x}-\boldsymbol{\mu},$
where $\boldsymbol{\mu}$ are the means. In particular, the TICA basis
set is linear in the input coordinates. Then the resulting estimates
of eigenfunctions can be viewed as a set of linearly independent components
(ICs) with autocorrelations $\lambda_{i}(\tau)$. The dominant ICs
can be used to reduce the dimension of the molecular system. Notice
that using mean free coordinates is equivalent to removing the stationary
spectral component $(\lambda_{1},\psi_{1})\equiv(1,\mathbbm1)$, thus
TICA will only contain the dynamical components, starting from $(\lambda_{2},\psi_{2})$.

In recent MD papers, the term TICA has also been used as the application
of Eqs. (\ref{eq:C0_eq_expectation}-\ref{eq:variational-eigenproblem})
on trajectories of features, such as distances, contact maps or angles,
i.e. the transformation $\boldsymbol{\chi}\left(\mathbf{x}_{t}\right)$
has been applied \cite{PerezEtAl_JCP13_TICA,SchwantesPande_JCTC13_TICA}.
In this paper we will avoid using the term TICA when VA is meant,
because a main result here is that in order to obtain a good variational
approximation of the spectral components in (\ref{eq:p-decomposition}),
it is necessary to employ specific estimation algorithms for (\ref{eq:C0_eq_expectation}-\ref{eq:Ct_eq_expectation})
that require the stationary spectral component $(\lambda_{1},\psi_{1})$
to be kept.

\noindent\textbf{MSM: }The MSM is a special case of the VA while
using the indicator functions as basis set:
\begin{equation}
\chi_{i}\left(\mathbf{x}\right)=\left\{ \begin{array}{ll}
1, & \text{for }\mathbf{x}\in A_{i},\\
0, & \text{for }\mathbf{x}\notin A_{i},
\end{array}\right.\label{eq:characteristic_basis_set}
\end{equation}
where $A_{1},\ldots,A_{m}$ form a partition of the phase space $\Omega$.
With such basis functions, the correlation matrix $\mathbf{C}\left(0\right)$
is a diagonal matrix with $\left[\mathbf{C}\left(0\right)\right]_{ii}=\Pr\left(\mathbf{x}_{t}\in A_{i}\right)$
being the equilibrium probability of $A_{i}$, and the $(i,j)$-th
element $\left[\mathbf{C}\left(\tau\right)\right]_{ij}=\Pr\left(\mathbf{x}_{t}\in A_{i},\mathbf{x}_{t+\tau}\in A_{j}\right)$
of the time-lagged correlation matrix $\mathbf{C}\left(\tau\right)$
is equal to the equilibrium frequency of the transition from $A_{i}$
to $A_{j}$. Thus, a piecewise-constant approximation of eigenfunctions
\begin{equation}
\psi_{j}\left(\mathbf{x}\right)=b_{ij},\text{ for }\mbox{\ensuremath{\mathbf{x}}}\in A_{i},
\end{equation}
and the corresponding eigenvalues are given by the generalized eigenvalue
problem (\ref{eq:variational-eigenproblem}). When the equilibrium
probability of each $A_{i}$ is positive, this problem can be equivalently
transformed into a simple eigenvalue problem by
\begin{equation}
\mathbf{C}\left(\tau\right)\mathbf{B}=\mathbf{C}\left(0\right)\mathbf{B}\boldsymbol{\Lambda}\quad\Rightarrow\mathbf{\quad P}\left(\tau\right)\mathbf{B}=\mathbf{B}\boldsymbol{\Lambda}.
\end{equation}
Here, $\mathbf{P}\left(\tau\right)=\mathbf{C}\left(0\right)^{-1}\mathbf{C}\left(\tau\right)$
is the transition matrix of the MSM with $\left[\mathbf{P}\left(\tau\right)\right]_{ij}=\Pr\left(\mathbf{x}_{t+\tau}\in A_{j}|\mathbf{x}_{t}\in A_{i}\right)$,
and is the Koopman model for the basis set ((\ref{eq:characteristic_basis_set})).
The viewpoint that MSMs can be viewed as an approximation to molecular
kinetics via a projection of eigenfunctions to a basis of characteristic
functions has been proposed earlier \cite{SarichNoeSchuette_MMS09_MSMerror}.

The choice of more general basis functions for the VA is beyond the
scope of this paper, and some related work can be found in \cite{NueskeEtAl_JCTC14_Variational,VitaliniNoeKeller_JCTC15_BasisSet,NueskeEtAl_JCP15_Tensor}. 

\subsection{Estimation of covariance matrices}

\label{subsec:estimation_covar}

The remaining problem is how to obtain estimates of $\mathbf{C}\left(0\right)$
and $\mathbf{C}\left(\tau\right)$. For convenience of notation, we
take all sampled coordinates $\mathbf{x}_{t}$ of a trajectory, evaluate
their basis function values $\boldsymbol{\chi}\left(\mathbf{x}_{t}\right)=\left(\chi_{1}\left(\mathbf{x}_{t}\right),\,...,\,\chi_{m}\left(\mathbf{x}_{t}\right)\right)^{\top}$,
and define the following two matrices of size $N\times m$:
\begin{align}
\mathbf{X}=\left(\begin{array}{ccc}
\chi_{1}\left(\mathbf{x}_{1}\right) & \cdots & \chi_{m}\left(\mathbf{x}_{1}\right)\\
\vdots &  & \vdots\\
\chi_{1}\left(\mathbf{x}_{T-\tau}\right) & \cdots & \chi_{m}\left(\mathbf{x}_{T-\tau}\right)
\end{array}\right) & \:\:\:\mathbf{Y}=\left(\begin{array}{ccc}
\chi_{1}\left(\mathbf{x}_{\tau+1}\right) & \cdots & \chi_{m}\left(\mathbf{x}_{\tau+1}\right)\\
\vdots &  & \vdots\\
\chi_{1}\left(\mathbf{x}_{T}\right) & \cdots & \chi_{m}\left(\mathbf{x}_{T}\right)
\end{array}\right),\label{eq:basis_transform}
\end{align}
where each row corresponds to one stored time-step. Thus, $\mathbf{X}$
contains the first $N=T-\tau$ time steps and $\mathbf{Y}$ contains
the last $N=T-\tau$ time steps. Assuming that $\{\mathbf{x}_{t}\}$
is ergodic, $\mathbf{C}\left(0\right)$ and $\mathbf{C}\left(\tau\right)$
can be directly estimated by time averages of $\boldsymbol{\chi}\left(\mathbf{x}_{t}\right)\boldsymbol{\chi}\left(\mathbf{x}_{t}\right)^{\top}$
and $\boldsymbol{\chi}\left(\mathbf{x}_{t}\right)\boldsymbol{\chi}\left(\mathbf{x}_{t+\tau}\right)^{\top}$
over the trajectory:
\begin{eqnarray}
\hat{\mathbf{C}}\left(0\right) & = & \frac{1}{N}\mathbf{X}^{\top}\mathbf{X},\label{eq:C-hat-0}\\
\hat{\mathbf{C}}\left(\tau\right) & = & \frac{1}{N}\mathbf{X}^{\top}\mathbf{Y.}\label{eq:C-hat-1}
\end{eqnarray}
Furthermore, multiple trajectories $k=1,\,...,\,K$ are trivially
handled by adding up their contributions, e.g. $\hat{\mathbf{C}}\left(0\right)=\frac{1}{\sum_{k}N_{k}}\sum_{k}\mathbf{X}_{k}^{\top}\mathbf{X}_{k}$,
etc. For all covariance estimates in this paper we can employ the
shrinkage approach \cite{ledoit2004well,schafer2005shrinkage} in
order to reduce the sensitivity of estimated covariances to statistical
noise \cite{james2014concise} and improve the robustness of eigenvalues
and eigenvectors computed from (\ref{eq:C-hat-0}-\ref{eq:C-hat-1}).

Due to statistical noise or non-equilibrium starting points, the time-lagged
covariance matrix $\hat{\mathbf{C}}\left(\tau\right)$ estimated by
this method is generally not symmetric, even if the underlying dynamics
are time-reversible. Thus, the eigenvalue problem (\ref{eq:variational-eigenproblem})
may yield complex eigenvalues and eigenvectors, which are undesirable
in analysis of statistically reversible MD simulations. The relaxation
timescales $t_{i}$ can be computed from complex eigenvalues as $t_{i}=-\tau/\ln\left|\lambda_{i}\left(\tau\right)\right|$
by using the norm of eigenvalues, but it is \emph{a priori} unclear
how to perform component analysis and dimension reduction as in TICA
based on complex eigenfunctions.

In order to avoid the problem of complex estimates, a symmetric estimator
is often used in applications, which approximates $\mathbf{C}\left(0\right)$
and $\mathbf{C}\left(\tau\right)$ by empirically averaging over all
transition pairs $(\mathbf{x}_{t},\mathbf{x}_{t+\tau})$ and their
reverses $(\mathbf{x}_{t+\tau},\mathbf{x}_{t})$, which is equivalent
to averaging the time-forward and the time-inverted trajectory:
\begin{eqnarray}
\hat{\mathbf{C}}_{\mathrm{sym}}\left(0\right) & \approx & \frac{1}{2N}\left(\mathbf{X}^{\top}\mathbf{X}+\mathbf{Y}^{\top}\mathbf{Y}\right),\label{eq:va-C-hat-0}\\
\hat{\mathbf{C}}_{\mathrm{sym}}\left(\tau\right) & \approx & \frac{1}{2N}\left(\mathbf{X}^{\top}\mathbf{Y}+\mathbf{Y}^{\top}\mathbf{X}\right),\label{eq:va-C-hat-1}
\end{eqnarray}
so that the estimate of $\mathbf{C}\left(\tau\right)$ is always symmetric
and the generalized eigenvalue problem (\ref{eq:variational-eigenproblem})
has real-valued solutions.

For equilibrium simulations, i.e. if the simulation starting points
are sampled from the global equilibrium, or the simulations are much
longer than the slowest relaxation times, Eqs.~(\ref{eq:va-C-hat-0})
and (\ref{eq:va-C-hat-1}) are unbiased estimates of $\mathbf{C}_{\mu}\left(0\right)$
and $\mathbf{C}_{\mu}\left(\tau\right)$ and can also be derived from
the maximum likelihood estimation by assuming a multivariate normal
distribution of $(\mathbf{x}_{t},\mathbf{x}_{t+\tau})$ \cite{SchwantesPande_JCTC15_kTICA}.
The major difficulty of this approach arises from non-equilibrium
data, i.e. simulations whose starting points are not drawn from the
equilibrium distribution and are not long enough to reach that equilibrium
during the simulation. In this situation, (\ref{eq:va-C-hat-0}) and
(\ref{eq:va-C-hat-1}) do not converge to the true covariance matrices
in the limit of infinitely many trajectories, and may thus provide
biased estimates of the eigenvalues and eigenfunctions or independent
components.

The difference between the direct estimation and symmetric estimation
methods of covariance matrices becomes clear when considering the
MSM special case. Since the transition matrix is $\mathbf{P}=\mathbf{C}(0)^{-1}\mathbf{C}(\tau)$,
as shown in Section \ref{subsec:Linear-variational-approach}, transition
matrices of MSMs given by the two estimators are
\begin{align}
\left[\mathbf{P}\right]_{ij}= & \frac{c_{ij}(\tau)}{\sum_{k=1}^{m}c_{ik}(\tau)}, & \text{(direct estimation)}\\
\left[\mathbf{P}\right]_{ij}= & \frac{c_{ij}(\tau)+c_{ji}(\tau)}{\sum_{k=1}^{m}c_{ik}(\tau)+c_{ki}(\tau)}, & \text{(symmetric estimation)}
\end{align}
respectively. If the transition dynamics between discrete states $A_{1},\ldots,A_{m}$
are exactly Markovian, the direct estimator converges to the true
transition matrix in the large-data limit for non-equilibrium or even
nonreversible simulations, whereas the symmetric estimator does not.
However, the direct estimator may give a nonreversible transition
matrix with complex eigenvalues, which is why the symmetric estimator
has been frequently used before 2008 until it has been replaced by
reversible maximum likelihood and Bayesian estimators \cite{Noe_JCP08_TSampling,BucheteHummer_JPCB08,Bowman_JCP09_Villin,PrinzEtAl_JCP10_MSM1,TrendelkampNoe_JCP13_EfficientSampler,TrendelkampSchroerEtAl_InPrep_revMSM}.
How do we resolve this problem in the more general case of variational
estimates with arbitrary basis functions $\boldsymbol{\chi}$? Below,
we will introduce a solution based on Koopman operator theory and
dynamic mode decomposition (DMD).

\section{Koopman models of equilibrium kinetics}

A method equivalent to the linear VA algorithm described in \cite{NoeNueske_MMS13_VariationalApproach,NueskeEtAl_JCTC14_Variational}
and summarized in Sec. \ref{subsec:Linear-variational-approach} has
more recently been introduced in the fluid mechanics field under the
name extended Dynamic Mode Decomposition (EDMD) \cite{williams2015data}.
EDMD also projects the data onto basis functions, and approximates
the same eigenvalue and eigenfunctions like the linear VA. EDMD was
developed independently of the VA and is based on Dynamic Mode Decomposition
(DMD) \cite{SchmidSesterhenn_APS08_DMD,Schmid_JFM10_DMD,TuEtAl_JCD14_ExactDMD}.
EDMD and DMD approximate components of the Koopman operator which
is a generalization of the backward operator usually used in molecular
kinetics \cite{RowleyEtAl_JFM09_DMDSpectral,KlusKoltaiSchuette_ApproximationKoopman}.

The equivalence between the VA and EDMD is striking, because the EDMD
algorithm has been derived in a setting where dynamics are not reversible
and may not even possess a unique stationary distribution. In practice,
the EDMD algorithm effectively performs non-reversible empirical estimates
of the covariances (\ref{eq:C-hat-0}-\ref{eq:C-hat-1}) and is used
in non-equilibrium situations. EDMD is thus used in a regime for which
the variational principle does not hold, and yet it does make a useful
approximation to eigenvalue and eigenfunctions of dynamical operators
\cite{williams2015data}. This has two important consequences:
\begin{enumerate}
\item The linear VA is also usable for systems or data that are not reversible
and not in equilibrium.
\item We can use ideas from EDMD and Koopman operator theory to obtain equilibrium
and reversible estimates from non-equilibrium, non-reversible data.
\end{enumerate}
In this section we will develop the second point and construct estimators
for equilibrium expectations from non-equilibrium data. This will
allow us to estimate relaxation timescales, slow collective variables
and equilibrium expectations using arbitrary basis sets and without
requiring a cluster discretization as used in MSMs.

\subsection{Koopman operator description of conformation dynamics\label{subsec:Koopman-Ooperator-description}}

According to the Koopman operator theory \cite{mezic2005spectral},
the dynamics of a Markov process $\{\mathbf{x}_{t}\}$ can be fully
described by an integral operator $\mathcal{K}_{\tau}$, called \emph{Koopman
operator}, which maps an observable quantity $f\left(\mathbf{x}_{t}\right)$
at time $t$, to its expectation at time $t+\tau$ as
\begin{eqnarray}
\mathcal{K}_{\tau}f\left(\mathbf{x}\right) & = & \mathbb{E}\left[f\left(\mathbf{x}_{t+\tau}\right)|\mathbf{x}_{t}=\mathbf{x}\right]\nonumber \\
 & = & \int\mathrm{d}\mathbf{y}\ p\left(\mathbf{x},\mathbf{y};\tau\right)f\left(\mathbf{y}\right).
\end{eqnarray}
If the dynamics fulfill detailed balance, the spectral components
$\{(\lambda_{i},\psi_{i})\}$ discussed in Section \ref{sec:Variational-Approach}
are in fact the eigenvalues and eigenfunctions of the Koopman operator:
\begin{equation}
\mathcal{K}_{\tau}\psi_{i}=\lambda_{i}\psi_{i}\label{eq:koopman-eigenproblem}
\end{equation}
Notice that the operator description and decomposition of molecular
kinetics can also be equivalently provided by the forward and backward
operators, which propagate ensemble densities instead of observables
\cite{PrinzEtAl_JCP10_MSM1}. We exploit the Koopman operator in this
paper because it is the only one of these operators that can be reliably
approximated from non-equilibrium data in general. See Section \ref{subsec:edmd}
and Appendix \ref{sec:Dynamical-operators} for a more detailed analysis.

Eq.~(\ref{eq:koopman-eigenproblem}) suggests the following way for
spectral estimation: First approximate the Koopman operator from data,
and then extract the spectral components from the estimated operator.

\subsection{Using linear VA for non-equilibrium and non-reversible data: Extended
dynamic mode decomposition\label{subsec:edmd}}

Like in the VA, we can also approximate the Koopman operator $\mathcal{K}_{\tau}$
by its projection $\mathcal{K}_{\tau}^{\mathrm{proj}}$ onto the subspace
spanned by basis functions $\boldsymbol{\chi}$ which satisfies
\begin{equation}
\mathcal{K}_{\tau}f\approx\mathcal{K}_{\tau}^{\mathrm{proj}}f\in\mathrm{span}\{\chi_{1},\ldots,\chi_{m}\}\label{eq:Koopman-approximation}
\end{equation}
for any function $f$ in that subspace. As the Koopman operator is
linear, even if the dynamics are nonlinear, it can be approximated
by a matrix $\mathbf{K}=(\mathbf{k}_{1},\ldots,\mathbf{k}_{m})\in\mathbb{R}^{m\times m}$
as
\begin{equation}
\mathcal{K}_{\tau}^{\mathrm{proj}}\left(\sum_{i=1}^{m}c_{i}\chi_{i}\right)=\sum_{i=1}^{m}c_{i}\mathbf{k}_{i}^{\top}\boldsymbol{\chi},
\end{equation}
with
\begin{equation}
\mathbf{k}_{i}^{\top}\boldsymbol{\chi}=\mathcal{K}_{\tau}^{\mathrm{proj}}\chi_{i}\approx\mathcal{K}_{\tau}\chi_{i}
\end{equation}
representing a finite-dimensional approximation of $\mathcal{K}_{\tau}\chi_{i}$.
After a few algebraic steps \cite{williams2015data}, it can be shown
that eigenfunctions of $\mathcal{K}_{\tau}^{\mathrm{proj}}$ also
have the form $\psi_{i}=\mathbf{b}_{i}^{\top}\boldsymbol{\chi}$,
and eigenvalues and eigenfunctions of $\mathcal{K}_{\tau}^{\mathrm{proj}}$
can be calculated by the eigenvalue problem
\begin{equation}
\mathbf{K}\mathbf{B}=\mathbf{B}\boldsymbol{\Lambda},\label{eq:K-eigenproblem}
\end{equation}
where definitions of $\boldsymbol{\Lambda},\mathbf{B}$ are the same
as in (\ref{eq:variational-eigenproblem}). Considering that
\begin{equation}
\mathbb{E}\left[\chi_{i}\left(\mathbf{x}_{t+\tau}\right)|\mathbf{x}_{t}\right]=\mathcal{K}_{\tau}\chi_{i}\left(\mathbf{x}_{t}\right)\approx\mathbf{k}_{i}^{\top}\boldsymbol{\chi}\left(\mathbf{x}_{t}\right)
\end{equation}
for each transition pair $(\mathbf{x}_{t},\mathbf{x}_{t+\tau})$ in
simulations, the matrix $\mathbf{K}$ can be determined via minimizing
the mean square error between $\mathbf{k}_{i}^{\top}\boldsymbol{\chi}\left(\mathbf{x}_{t}\right)$
and $\chi_{i}\left(\mathbf{x}_{t+\tau}\right)$:
\begin{eqnarray}
\mathbf{K} & = & \arg\min_{\mathbf{K}}\frac{1}{N}\sum_{t=1}^{T-\tau}\sum_{i=1}^{m}\left\Vert \mathbf{k}_{i}^{\top}\boldsymbol{\chi}\left(\mathbf{x}_{t}\right)-\chi_{i}\left(\mathbf{x}_{t+\tau}\right)\right\Vert ^{2}\nonumber \\
 & = & \arg\min_{\mathbf{K}}\frac{1}{N}\left\Vert \mathbf{X}\mathbf{K}-\mathbf{Y}\right\Vert ^{2}\nonumber \\
 & = & \hat{\mathbf{C}}\left(0\right)^{-1}\hat{\mathbf{C}}\left(\tau\right).\label{eq:K-approximation}
\end{eqnarray}
where the covariance matrices are given by their direct estimates
(\ref{eq:C-hat-0}-\ref{eq:C-hat-1}), and $\left\Vert \cdot\right\Vert $
denotes the Frobenius norm of matrices. 

Based on the above considerations it makes sense to call the matrix
$\mathbf{K}$ together with the basis set $\boldsymbol{\chi}$ a \emph{Koopman
model} of the molecular kinetics. The Koopman model is a generalization
of an MSM, as it can be constructed from any basis set $\boldsymbol{\chi}$,
not only from characteristic basis sets (Eq. (\ref{eq:characteristic_basis_set})).
Nonetheless, it shares the main features of an MSM as it can be used
to propagate densities according to ((\ref{eq:Koopman-approximation})),
and its eigenvalues can be used to compute relaxation timescales and
its eigenvectors can be used to compute slow collective variables.
The following algorithm computes a nonreversible Koopman model from
data. This algorithm is equivalent to the linear VA and EDMD. If the
feature trajectories are mean-free, it is also equivalent to TICA
in feature space:

\noindent\textbf{Algorithm 1: Nonreversible Koopman estimation}
\begin{enumerate}
\item Basis-transform input coordinates according to (\ref{eq:basis_transform}).
\item Compute $\hat{\mathbf{C}}\left(0\right)=\frac{1}{N}\mathbf{X}^{\top}\mathbf{X}$
and $\hat{\mathbf{C}}\left(\tau\right)=\frac{1}{N}\mathbf{X}^{\top}\mathbf{Y}$.
\item Compute the Koopman model $\mathbf{K}=\hat{\mathbf{C}}\left(0\right)^{-1}\hat{\mathbf{C}}\left(\tau\right)$.
\item Koopman decomposition: Solve eigenvalue problem $\mathbf{K}\mathbf{B}=\mathbf{B}\boldsymbol{\Lambda}$.
\item Output the Koopman model $\mathbf{K}$ and spectral components: Eigenvalues
$\lambda_{i}$ and eigenfunctions $\psi_{i}=\mathbf{b}_{i}^{\top}\boldsymbol{\chi}$.
Both may have imaginary components that are either due to statistical
noise or nonreversible dynamics.
\end{enumerate}
Please note that this pseudocode is given for illustrative purposes
and should not be implemented literally. In particular, it assumes
that the basis functions are linearly independent so that $\hat{\mathbf{C}}\left(0\right)$
is invertible. In practice, linear independence can be achieved by
de-correlation of basis functions - see Appendix \ref{sec:Detailed-decorreltion-procedure}
and specifically Algorithm 1{*} there for advice how to practically
implement the Koopman estimator.

The above derivation shows that Koopman estimation as in Algorithm
1 has a key advantage: Suppose the points $\{\mathbf{x}_{1},\ldots,\mathbf{x}_{T-\tau}\}$
are sampled from a distribution $\rho\left(\mathbf{x}\right)$ which
is not equal to the equilibrium distribution $\mu$. Although the
empirical estimates of covariance matrices used in Algorithm 1 are
biased with respect to the equilibrium expectations $\mathbf{C}\left(0\right)$
and $\mathbf{C}\left(\tau\right)$, they are unbiased and consistent
estimates of the non-equilibrium covariance matrices $\mathbb{E}_{\rho}\left[\boldsymbol{\chi}\left(\mathbf{x}_{t}\right)\boldsymbol{\chi}\left(\mathbf{x}_{t}\right)^{\top}\right]$
and $\mathbb{E}_{\rho}\left[\boldsymbol{\chi}\left(\mathbf{x}_{t}\right)\boldsymbol{\chi}\left(\mathbf{x}_{t+\tau}\right)^{\top}\right]$.
Furthermore, the matrix $\mathbf{K}$ given by (\ref{eq:K-approximation})
minimizes the error
\begin{equation}
\sum_{i}\left\langle \mathbf{k}_{i}^{\top}\boldsymbol{\chi}-\mathcal{K}_{\tau}\chi_{i},\mathbf{k}_{i}^{\top}\boldsymbol{\chi}-\mathcal{K}_{\tau}\chi_{i}\right\rangle _{\rho},
\end{equation}
with $\left\langle f,g\right\rangle _{\rho}\triangleq\int\mathrm{d}\mathbf{x}\ \rho\left(\mathbf{x}\right)f\left(\mathbf{x}\right)g\left(\mathbf{x}\right)$,
as data size approaches infinity (see Appendices \ref{sec:Properties-of-rho}
and \ref{sec:Limit-of-the-error}). Therefore, $\mathbf{K}$ is still
a finite-dimensional approximation of $\mathcal{K}_{\tau}$ with minimal
mean square error with respect to $\rho$, which implies that Algorithm
1 is applicable to non-equilibrium data.

\subsection{Koopman reweighting: Estimation the equilibrium distribution from
non-equilibrium data}

Not only is EDMD robust when using non-equilibrium data, we can also
utilize the Koopman matrix $\mathbf{K}$ to recover the equilibrium
properties of the molecular system. According to the principle of
importance sampling \cite{glynn1989importance}, we can assign a weight
\begin{equation}
w(\mathbf{x}_{t})\propto\frac{\mu(\mathbf{x}_{t})}{\rho(\mathbf{x}_{t})}\label{eq:importance_weight}
\end{equation}
to each transition pair $(\mathbf{x}_{t},\mathbf{x}_{t+\tau})$ in
the simulation data, such that the equilibrium ensemble average of
a function $h\left(\mathbf{x}_{t},\mathbf{x}_{t+\tau}\right)$ can
be consistently estimated by the weighted mean as
\begin{equation}
\mathbb{E}_{\mu}\left[h\left(\mathbf{x}_{t},\mathbf{x}_{t+\tau}\right)\right]\approx\frac{\sum_{t=1}^{T-\tau}w\left(\mathbf{x}_{t}\right)h\left(\mathbf{x}_{t},\mathbf{x}_{t+\tau}\right)}{\sum_{t=1}^{T-\tau}w\left(\mathbf{x}_{t}\right)}.\label{eq:reweighting}
\end{equation}
Based on the finite-dimensional approximation (\ref{eq:eigenfunction-approximation})
of spectral components, we can represent the weight function as a
linear combination of basis functions $\boldsymbol{\chi}$:
\begin{equation}
w\left(\mathbf{x}\right)=\mathbf{u}^{\top}\boldsymbol{\chi}\left(\mathbf{x}\right),\label{eq:Koopman-reweighting}
\end{equation}
where $\mathbf{u}$ satisfies
\begin{equation}
\hat{\mathbf{C}}\left(0\right)^{-1}\mathbf{K}^{\top}\hat{\mathbf{C}}\left(0\right)\mathbf{u}=\mathbf{u}\label{eq:u-direction}
\end{equation}
i.e., $\mathbf{u}$ is the eigenvector of $\hat{\mathbf{C}}\left(0\right)^{-1}\mathbf{K}^{\top}\hat{\mathbf{C}}\left(0\right)$
with eigenvalue $1$, in the limit of large statistics. (See Appendix
\ref{sec:Properties-of-rho} for proofs of the above equations.)

In practice, the eigenvalue problem (\ref{eq:u-direction}) cannot
be solved for arbitrary choices of basis sets. If the basis set cannot
represent the constant 1 eigenfunction, (\ref{eq:u-direction}) does
not have an eigenvalue 1. In order to deal with general basis sets,
we have two options. First, we can seek an approximate solution \emph{via}
the quadratic programming problem
\begin{equation}
\begin{array}{rl}
\min_{\mathbf{u}} & \left\Vert \hat{\mathbf{C}}\left(0\right)^{-1}\mathbf{K}^{\top}\hat{\mathbf{C}}\left(0\right)\mathbf{u}-\mathbf{u}\right\Vert ^{2}\\
\mathrm{s.t.} & \mathbf{1}^{\top}\mathbf{X}\mathbf{u}=1
\end{array}\label{eq:u-quadratic-programming}
\end{equation}
where the constraint $\mathbf{1}^{\top}\mathbf{X}\mathbf{u}$ ensures
that $\sum_{t=1}^{T-\tau}w\left(\mathbf{x}_{t}\right)=1$, and $\mathbf{1}$
denotes a column vector of ones. 

We recommend a simpler way to solve this problem: Add the constant
function $\mathbbm1$ to the basis set and change $\mathbf{X}$ and
$\mathbf{Y}$ as $\mathbf{X}:=[\mathbf{X}\;\mathbf{1}]$ and $\mathbf{Y}:=[\mathbf{Y}\;\mathbf{1}]$
correspondingly so that the eigenvalue problem (\ref{eq:u-quadratic-programming})
can be exactly solved. The resulting method to compute equilibrium
statistical weights of all samples, $w\left(\mathbf{x}_{t}\right)$,
can be summarized by the following algorithm:

\noindent\textbf{Algorithm 2: Koopman reweighting}
\begin{enumerate}
\item Basis-transform input coordinates according to (\ref{eq:basis_transform}).
\item Compute $\hat{\mathbf{C}}\left(0\right)=\frac{1}{N}\mathbf{X}^{\top}\mathbf{X}$,
$\hat{\mathbf{C}}\left(\tau\right)=\frac{1}{N}\mathbf{X}^{\top}\mathbf{Y}$
and $\mathbf{K}=\hat{\mathbf{C}}\left(0\right)^{-1}\hat{\mathbf{C}}\left(\tau\right)$
as in Algorithm 1.
\item Compute $\mathbf{u}$ as eigenvector of $\hat{\mathbf{C}}\left(0\right)^{-1}\mathbf{K}^{\top}\hat{\mathbf{C}}\left(0\right)$
with eigenvalue 1, and normalize it by $\mathbf{1}^{\top}\mathbf{X}\mathbf{u}$.
\item Output weights: $w\left(\mathbf{x}_{t}\right)=\mathbf{x}_{t}^{\top}\mathbf{u}$.
\end{enumerate}
Again, this algorithm is simplified for illustrative purposes. In
our implementation, we ensure numerical robustness by adding the constant
function to the decorrelated basis set - see Appendix \ref{sec:Analysis-of-reversibility}
and Algorithm 2{*} there.

After the weights $w\left(\mathbf{x}_{t}\right)$ have been estimated
and normalized with $\sum_{t=1}^{T-\tau}w\left(\mathbf{x}_{t}\right)=\mathbf{1}^{\top}\mathbf{X}\mathbf{u}=1$,
we can compute equilibrium estimates for given observables $f_{1}\left(\mathbf{x}_{t}\right)$
and $f_{2}\left(\mathbf{x}_{t}\right)$ from non-equilibrium data.
For example, ensemble average and time-lagged cross correlation can
be approximated by
\begin{eqnarray}
\mathbb{E}_{\mu}\left[f_{1}\left(\mathbf{x}_{t}\right)\right] & \approx & \sum_{t=1}^{T-\tau}w\left(\mathbf{x}_{t}\right)\cdot f_{1}(\mathbf{x}_{t}),\label{eq:eq_f1}\\
\mathbb{E}_{\mu}\left[f_{1}\left(\mathbf{x}_{t}\right)f_{2}\left(\mathbf{x}_{t+\tau}\right)\right] & \approx & \sum_{t=1}^{T-\tau}w\left(\mathbf{x}_{t}\right)\cdot f_{1}\left(\mathbf{x}_{t}\right)f_{2}\left(\mathbf{x}_{t+\tau}\right).\label{eq:eq_f1f2}
\end{eqnarray}

\subsection{Reversible Koopman models and Eigendecompositions\label{subsec:Reversible-EDMD}}

We now have the tools necessary to compute equilibrium covariance
matrices while avoided the bias of forced symmetrization described
in Sec. \ref{subsec:estimation_covar}, and can conduct real-valued
eigenvalue analysis for reversible dynamics using VA or TICA. At the
same time our approach defines an equilibrium estimator of EDMD for
time-reversible processes. We can obtain symmetrized equilibrium covariances
from our off-equilibrium data by the following estimators:

\begin{eqnarray}
\hat{\mathbf{C}}_{\mathrm{rev}}\left(0\right) & = & \frac{1}{2}\sum_{t=1}^{T-\tau}w\left(\mathbf{x}_{t}\right)\left(\boldsymbol{\chi}(\mathbf{x}_{t})\boldsymbol{\chi}(\mathbf{x}_{t})^{\top}+\boldsymbol{\chi}(\mathbf{x}_{t+\tau})\boldsymbol{\chi}(\mathbf{x}_{t+\tau})^{\top}\right)\nonumber \\
 & = & \frac{1}{2}\left(\mathbf{X}^{\top}\mathbf{W}\mathbf{X}+\mathbf{Y}^{\top}\mathbf{W}\mathbf{Y}\right)\label{eq:Sigma}\\
\hat{\mathbf{C}}_{\mathrm{rev}}\left(\tau\right) & = & \frac{1}{2}\sum_{t=1}^{T-\tau}w\left(\mathbf{x}_{t}\right)\left(\boldsymbol{\chi}(\mathbf{x}_{t})\boldsymbol{\chi}(\mathbf{x}_{t+\tau})^{\top}+\boldsymbol{\chi}(\mathbf{x}_{t+\tau})\boldsymbol{\chi}(\mathbf{x}_{t})^{\top}\right)\nonumber \\
 & = & \frac{1}{2}\left(\mathbf{X}^{\top}\mathbf{W}\mathbf{Y}+\mathbf{Y}^{\top}\mathbf{W}\mathbf{X}\right).\label{eq:edmd-symmetrization}
\end{eqnarray}
These estimators are based on time-reversibility and the reweighting
approximation (\ref{eq:reweighting}) for the equilibrium distribution
(see Appendix \ref{sec:Proof-of-symmetrization} for proof). As a
result, we obtain a time-reversible Koopman matrix:
\begin{equation}
\mathbf{K}_{\mathrm{rev}}=\hat{\mathbf{C}}_{\mathrm{rev}}\left(0\right)^{-1}\hat{\mathbf{C}}_{\mathrm{rev}}\left(\tau\right).\label{eq:reversibility-modification}
\end{equation}
By comparing (\ref{eq:Sigma},\ref{eq:edmd-symmetrization}) and (\ref{eq:va-C-hat-0},\ref{eq:va-C-hat-1}),
it is apparent that $\hat{\mathbf{C}}_{\mathrm{rev}}\left(0\right)$
and $\hat{\mathbf{C}}_{\mathrm{rev}}\left(\tau\right)$ are equal
to the symmetrized direct estimates if weights of data are uniform
with $\mathbf{W}=\mathrm{diag}\left(\frac{1}{N},\ldots,\frac{1}{N}\right)$.
The weight function (\ref{eq:Koopman-reweighting}) used here can
systematically reduce the bias of the symmetrized estimates for reversible
dynamics. Under some weak assumptions, it can be shown that the spectral
components calculated from $\mathbf{K}_{\mathrm{rev}}$ are real-valued
and the largest eigenvalue is not larger than $1$ even in the existence
of statistical noise and modeling error. Furthermore, the procedure
is self-consistent: If the estimation procedure is repeated while
starting with weights $w\left(\mathbf{x}\right)$, these weights remain
fixed (See Appendix \ref{sec:Analysis-of-reversibility} for more
detailed analysis.)

The estimation algorithm for variationally optimal Koopman models
of the reversible equilibrium dynamics can be summarized as follows:

\noindent\textbf{Algorithm 3: Reversible Koopman estimation}
\begin{enumerate}
\item Basis-transform input coordinates according to (\ref{eq:basis_transform}).
\item Use Koopman reweighting (Algorithm 2) to compute the equilibrium weights
$w\left(\mathbf{x}_{t}\right)$.
\item Compute $\hat{\mathbf{C}}_{\mathrm{rev}}\left(0\right)$ and $\hat{\mathbf{C}}_{\mathrm{rev}}\left(\tau\right)$
by (\ref{eq:Sigma}) and (\ref{eq:edmd-symmetrization}).
\item Compute the Koopman model $\mathbf{K}_{\mathrm{rev}}=\hat{\mathbf{C}}_{\mathrm{rev}}\left(0\right)^{-1}\hat{\mathbf{C}}_{\mathrm{rev}}\left(\tau\right)$.
\item Reversible Koopman decomposition: solve eigenvalue problem $\mathbf{K}_{\mathrm{rev}}\mathbf{B}=\mathbf{B}\boldsymbol{\Lambda}$.
\item Output the Koopman model $\mathbf{K}_{\mathrm{rev}}$ and its spectral
components: Eigenvalues $\lambda_{i}$ and eigenfunctions $\psi_{i}=\mathbf{b}_{i}^{\top}\boldsymbol{\chi}$.
These eigenvalues and eigenfunctions are real-valued.
\end{enumerate}
As before, this algorithm is presented in a pedagogical pseudocode.
Taken literally, it will suffer from numerical instabilities if $\hat{\mathbf{C}}_{\mathrm{rev}}\left(0\right)$
is not positive-definite, which can also be overcome by reducing correlations
between basis functions as mentioned in Section \ref{subsec:edmd}
- see Appendix \ref{sec:Detailed-decorreltion-procedure} and Algorithm
3{*} there.

\section{Applications}

In this section, we compare three different estimators for molecular
kinetics to the same data sets:
\begin{enumerate}
\item VA or TICA in feature space symmetrization of covariance matrices
(\ref{eq:va-C-hat-0}-\ref{eq:va-C-hat-1}), as proposed before \cite{SchwantesPande_JCTC13_TICA,PerezEtAl_JCP13_TICA}.
Briefly we refer to this estimator as symmetrized VA or symmetrized
TICA.
\item Nonreversible Koopman estimation (Algorithm 1), which provides a nonreversible
Koopman model whose eigendecomposition is equivalent to EDMD and (nonsymmetrized)
TICA in feature space.
\item Reversible Koopman estimation (Algorithm 3), which is consistent with
the variational approach \cite{NoeNueske_MMS13_VariationalApproach,NueskeEtAl_JCTC14_Variational}. 
\end{enumerate}
In addition, we compare the estimated equilibrium distribution provided
by Koopman reweighting (Algorithm 2) with the empirical distribution
estimated from direct counting or histogramming the data in order
to demonstrate the usefulness of the proposed reweighting method.

\subsection{One-dimensional diffusion process\label{subsec:One-dimensional-diffusion-process}}

\begin{figure}
\begin{centering}
\includegraphics[width=1\textwidth]{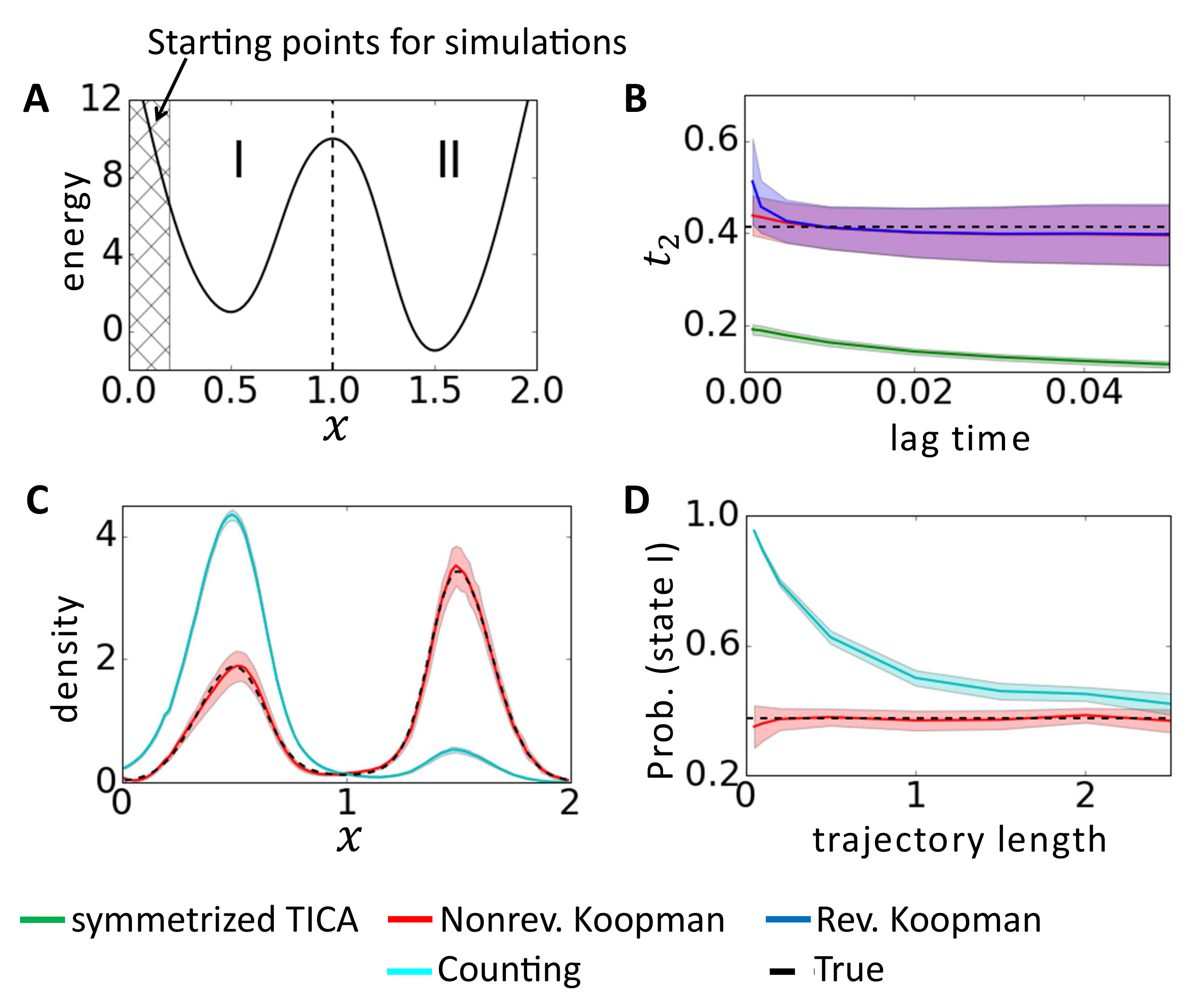}
\par\end{centering}
\caption{Estimation results of a one-dimensional diffusion process. (A) Dimensionless
energy $U\left(x\right)$, where the dashed line represents the border
of the two potential wells I and II. The shaded area denotes the region
where initial states are drawn for simulations. (B) The slowest relaxation
timescale estimated by the previously used symmetrized TICA, nonreversible
Koopman estimation (Algorithm 1) and reversible Koopman estimation
(Algorithm 3) with different lag times. (C) Stationary density of
states obtained from equilibrium probabilities of $100$ uniform bins,
where the probabilities are estimated using Koopman reweighting (Algorithm
2, red) direct counting. (D) Estimates of the equilibrium probability
of the potential well I given by direct counting and the Koopman reweighting
(red) with different simulation trajectory lengths. In (B-D), solid
lines and shaded regions indicate mean values and one standard derivation
error intervals obtained from $30$ independent experiments.\label{fig:one-dimensional-diffusion}}
\end{figure}

As a first example, we consider a one-dimensional diffusion process
$\{x_{t}\}$ in a double-well potential restricted to the $x$-range
$[0,2]$ as shown in Fig.~\ref{fig:one-dimensional-diffusion}A.
In order to validate the robustness of different estimators, we start
all simulations far from equilibrium, in the region $[0,0.2]$ (shaded
area in Fig.~\ref{fig:one-dimensional-diffusion}A). In order to
apply the algorithms discussed here, we choose a basis set of $100$
Gaussian functions with random parameters. For more details on the
simulation model and experimental setup, see Appendix \ref{subsec:One-dimensional-setup}.

Fig.~\ref{fig:one-dimensional-diffusion}B shows estimates of the
slowest relaxation timescale $\mathrm{ITS}_{2}$ based on $500$ independent
short simulation trajectories with length $0.2$ time units. The largest
relaxation timescale $t_{2}$ is computed from $\lambda_{2}$ as $t_{2}=-\tau/\ln\left|\lambda_{2}\left(\tau\right)\right|$
and is a constant independent of lag time according to (\ref{eq:p-decomposition}).
For such non-equilibrium data, the symmetrized VA significantly underestimates
the relaxation timescale for such non-equilibrium data and gives even
worse results with longer lag times. The Koopman models (both reversible
and nonreversible), on the other hand, converge quickly to the true
timescale before $\tau=0.01$ time units. The equilibrium distribution
distribution of $\{x_{t}\}$ computed from Algorithm 2 with lag time
$0.01$ is shown in Fig.~\ref{fig:one-dimensional-diffusion}C. In
contrast to the empirical histogram density given by direct counting,
the direct estimator effectively recovers the equilibrium property
of the process from non-equilibrium data.

Fig.~\ref{fig:one-dimensional-diffusion}D compares the empirical
probability of the potential well I (i.e. by direct counting of the
number of samples in well I) with the estimate from Koopman reweighting
(Algorithm 2), for different simulation trajectory lengths, where
the lag time is still $0.01$ time units and the accumulated simulation
time is kept fixed to be $100$. Due to the ergodicity of the process,
the empirical probability converges to the true value as the trajectory
length increases. The convergence rate, however, is very slow as shown
in Fig.~\ref{fig:one-dimensional-diffusion}D, and empirical probability
is close to the true value only for trajectories longer than $2$
time units. When using the reweighting method proposed here, the estimated
probability is robust with respect to changes in trajectory length,
and accurate even for very short trajectories.

\subsection{Two-dimensional diffusion process\label{subsec:Two-dimensional-diffusion-process}}

\begin{figure}
\begin{centering}
\includegraphics[width=1\textwidth]{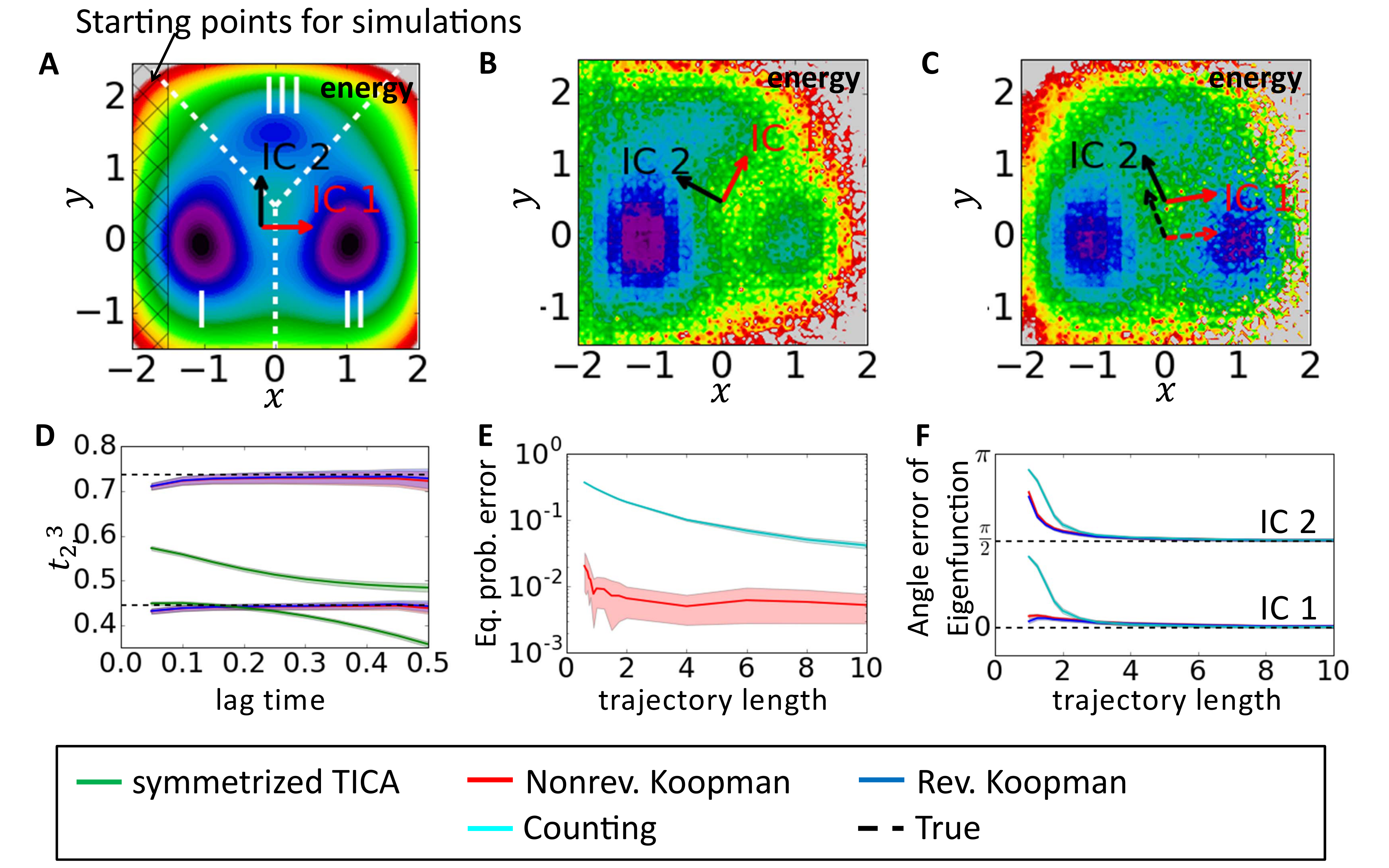}
\par\end{centering}
\caption{Estimation results of a two-dimensional diffusion process. (A) Free
energy of the process, where the dashed line represents the border
of potential wells I, II, and III. The shaded area denotes the region
where initial states are drawn for simulations, and the two linear
ICs obtained from TICA with exact statistics. (B) Free energies computed
from a histogram of the simulation data (direct counting). Arrows
show the directions of TICA components computed from symmetrized TICA.
(C) Free energies computed from Koopman reweighting (Algorithm 2).
Arrows show the directions of the slowest modes computed from a reversible
(solid arrows) and nonreversible (dashed arrows) Koopman estimation
using $\{x,y\}$ as basis set. (D) Estimates of the two slowest relaxation
timescales. (E) Estimation errors of equilibrium distributions using
direct counting or the Koopman model (red). (F) Error in the angles
of estimated eigenfunctions. Shaded area shows the standard deviation
computed from $30$ independent simulations.\label{fig:two-dimensional-diffusion}}
\end{figure}

Next, we study a two-dimensional diffusion process $\{(x_{t},y_{t})\}$
which has three potential wells as shown in Fig.~\ref{fig:two-dimensional-diffusion}A,
where all simulations are initialized with $(x_{0},y_{0})\in[-2,-1.5]\times[-1.5,2.5]$,
and the set of basis functions for spectral estimation consists of
$100$ Gaussian functions with random parameters (see Appendix \ref{subsec:Two-dimensional-setup}
for details).

We generate $8000$ short simulation trajectories with length $1.25$
and show the empirical free energy of the simulation data in Fig.~\ref{fig:two-dimensional-diffusion}B.
Comparing Fig.~\ref{fig:two-dimensional-diffusion}B and Fig.~\ref{fig:two-dimensional-diffusion}A,
it can be seen that most of the simulation data are distributed in
the area $x\le0$ and the empirical distribution of simulations is
very different from the equilibrium distribution. Therefore, eigenvalues/timescales
and eigenfunctions estimated by the symmetrized VA have large errors,
whereas the nonreversible and reversible Koopman model provide accurate
eigenvalues/timescales and eigenfunctions (Fig.~\ref{fig:two-dimensional-diffusion}D,F).
Moreover the equilibrium density can be recovered with high accuracy
using Koopman reweighting, although the data is far from equilibrium
(Fig.~\ref{fig:two-dimensional-diffusion}C). 

For such a two-dimensional process, it is also interesting to investigate
the slow collective variables predicted by TICA, i.e. directly using
the $x$ and $y$ coordinates as basis functions. Fig.~\ref{fig:two-dimensional-diffusion}A
displays the TICA components from the exact equilibrium distribution
with lag time $\tau=0.01$. Notice that the slowest mode is parallel
to x-axis, which is related to transitions between potential wells
I and II, and the second IC is parallel to the y-axis, which is related
to transitions between \{I,II\} and III. However, if we extract ICs
from simulation data by using TICA with symmetrized covariance matrices,
the result is significantly different as shown in Fig.~\ref{fig:two-dimensional-diffusion}B,
where the first IC characterizes transitions between I and III. The
ICs given by nonreversible and reversible Koopman models (Algorithms
1 and 3 here) can be seen in Fig.~\ref{fig:two-dimensional-diffusion}C.
They are still different from those in Fig.~\ref{fig:two-dimensional-diffusion}A
because the equilibrium distribution is difficult to approximate with
only linear basis functions, but much more accurate than the estimates
obtained by the previously used symmetric estimator in Fig.~\ref{fig:two-dimensional-diffusion}B.

Fig.~\ref{fig:two-dimensional-diffusion}E shows the estimation errors
of estimated equilibrium distribution obtained by using simulations
with different trajectory lengths, where the accumulated simulation
time is kept fixed to be $10^{4}$, the lag time for estimators is
$\tau=0.005$, and the error is evaluated as the total variation distance
between the estimated probability distributions of the three potential
wells and the true reference. Fig.~\ref{fig:two-dimensional-diffusion}F
shows angles of linear ICs approximated from the same simulation data
with lag time $\tau=0.01$. Both of the figures clearly demonstrate
the superiority of the Koopman models suggested here.

\subsection{Protein-Ligand Binding}

\begin{figure}
\begin{centering}
\includegraphics[width=1\textwidth]{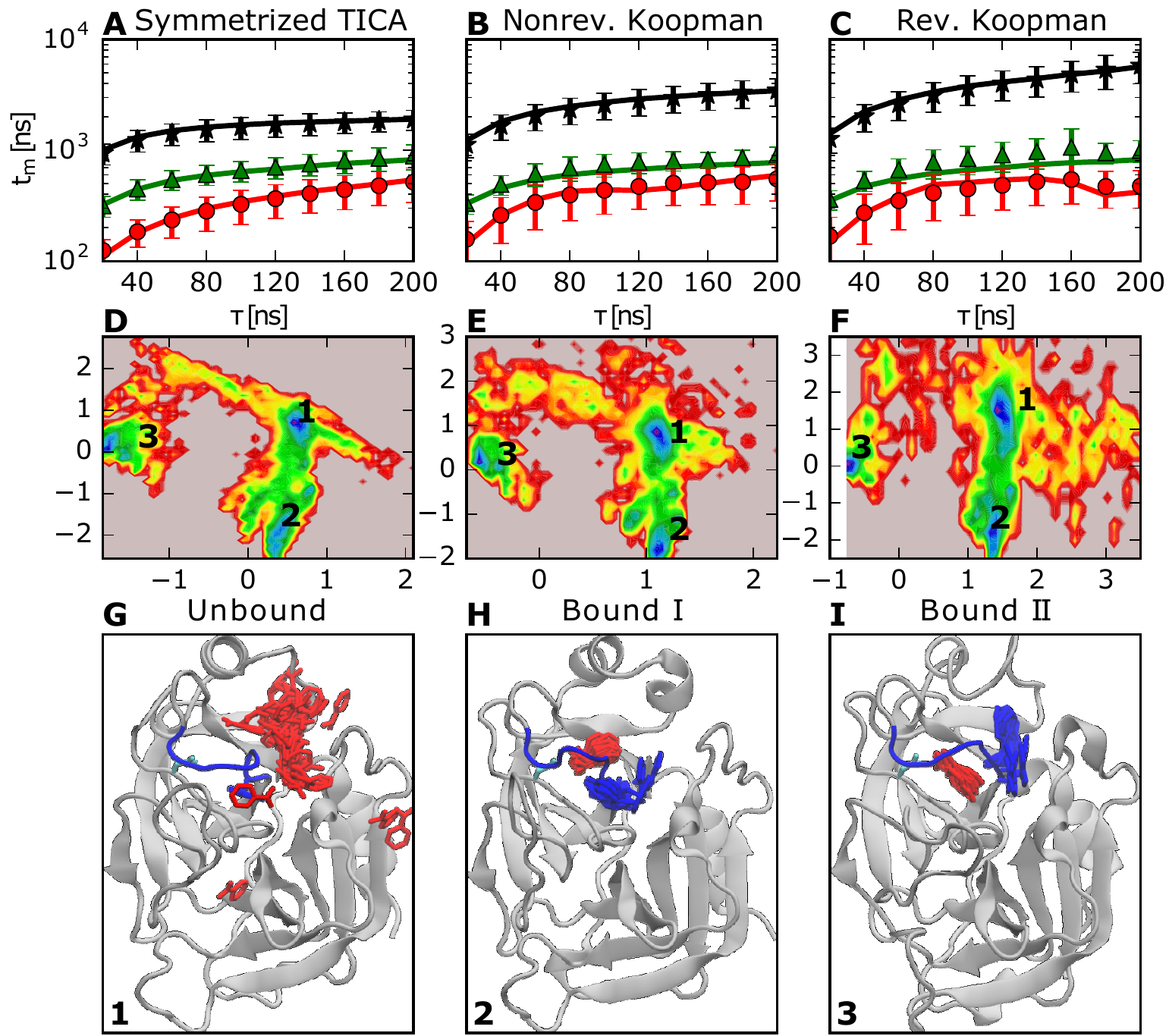}
\par\end{centering}
\caption{Results for MD simulations of the trypsin-benzamidine binding process.
\textbf{A-C}: Relaxation timescales are estimated as a function of
lag time. A: TICA in feature space with the previously used symmetric
estimator, B: Nonreversible Koopman model, equivalent to TICA in feature
space without symmetrization (Algorithm 1 here), C: Variational reversible
Koopman model suggested here (Algorithm 3). \textbf{D-I}: free energy
landscapes (negative logarithm of the sampled densities) plotted on
the two slowest process eigenfunctions. For all three methods, minima
1-3 correspond to the same macro-states of the system. Representative
structures of these states are shown in G-I. State 1 represents the
ligand being unbound or loosely attached to the protein. States 2
and 3 are different conformational arrangements of the bound state,
in particular of the binding loop including Trp 215 \cite{PlattnerNoe_NatComm15_TrypsinPlasticity}.\label{fig:Figure_Trp_Ben}}
\end{figure}

We revisit the the binding process of benzamidine to trypsin which
was studied previously in Refs. \cite{BuchFabritiis_PNAS11_Binding,SchererEtAl_JCTC15_EMMA2}.
The data set consists of $52$ trajectories of $2\mathrm{\mu s}$
and four trajectories of $1\mathrm{\,\mu s}$ simulation time, resulting
in a total simulation time of $108\,\mathrm{\mu s}$. From the simulations,
we extract a feature set of $223$ nearest neighbor heavy-atom contacts
between all trypsin residues and the ligand. In this feature space,
we then perform TICA using the symmetrized estimate (previous standard),
and estimate a nonreversible Koopman model (Algorithm 1) and a reversible
Koopman model (Algorithm 3). In order to obtain uncertainties, we
compute $100$ bootstrapping estimates in which outliers were rejected.
In Figure \ref{fig:Figure_Trp_Ben} A-C, we show the three slowest
implied timescales as estimated by the three approaches discussed
above. We observe that both the Koopman models provide a larger slowest
implied timescale than symmetrized TICA. The slowest timescale estimated
by the reversible estimator converges on relatively long lag times.
This is likely due to the fact that the trypsin-benzamidin binding
kinetics involves internal conformational changes of trypsin \cite{PlattnerNoe_NatComm15_TrypsinPlasticity}.
In Fig. \ref{fig:Figure_Trp_Ben} for all three estimates (the first
TICA components of the direct estimate are coincidentally purely real
here). The eigenvectors used for the dimensionality reduction were
estimated at lag time $\tau=100\,\mathrm{ns}$. The projections are
qualitatively similar, revealing three minima of the landscape, labeled
1, 2, and 3. In all three cases, these centers correspond to the same
macro-states of the system, shown underneath in Figure \ref{fig:Figure_Trp_Ben}
G-H. Center 1 corresponds to the ligand being either unbound or loosely
attached to the protein. The other two states are different conformational
arrangements of the bound state of the ligand. 

\section{Conclusion}

Using dynamic mode decomposition theory, we have shown that the variational
approach of conformation dynamics and the time-lagged independent
component analysis can be made with small bias even if just empirical
out-of-equilibrium estimates of the covariance matrices are available,
i.e. they can be applied to ensembles of short MD simulations starting
from arbitrary starting point. A crucial point is that the forceful
symmetrization of the empirical covariances practiced in previous
studies must be avoided. 

In order to facilitate a bias-corrected symmetric estimate of covariance
matrices, we have proposed a Koopman reweighting technique in which
the weights of sampled configurations can be estimated using a first
pass over the data, during which empirical covariance matrices must
be estimated. These weights can be applied in order to turn the empirical
(out-of-equilibrium) estimates of covariance matrices into estimates
of the equilibrium covariance matrices. These matrices can then be
symmetrized without introducing a bias from the empirical distribution,
resulting in real-valued eigenvalue and eigenfunction estimates.

With these algorithms, the variational approach, and thus also the
TICA algorithm inherit the same benefits that MSMs have enjoyed since
nearly a decade: we can generate optimal and robust reversible and
nonreversible estimates of the equilibrium kinetics from swarms of
short trajectories not started from equilibrium. Although this work
focuses on the estimation of eigenvalues and eigenfunctions of the
Koopman operator, the proposed Algorithms 1 and 3 provide Koopman
models, which are discrete approximations of the Koopman operator,
and that be used for other purposes, such as the propagation of densities.
Koopman models are generalization of Markov state models using arbitrary
basis functions.

Besides the application to molecular kinetics highlighted in this
paper, the Koopman reweighting principle described in Algorithm 2
can be used to compute variationally optimal estimates of any equilibrium
property (expectation values, distributions) from out-of-equilibrium
data using an approach that involves arbitrary sets of basis functions.
While the viability of this approach critically depends on the suitability
of the basis functions employed, it offers a very general way to computing
equilibrium expectations that may lead to other applications and extensions
in future studies.

\appendix

\section{Dynamical operators\label{sec:Dynamical-operators}}

Besides the Koopman operator $\mathcal{K}_{\tau}$, the conformation
dynamics of a molecular system can also be described by the forward
operator $\mathcal{P}_{\tau}$ and backward operator, or called transfer
operator, $\mathcal{T}_{\tau}$ \cite{PrinzEtAl_JCP10_MSM1}, which
describe the evolution of ensemble densities as
\begin{eqnarray}
p_{t+\tau}\left(\mathbf{x}\right) & = & \mathcal{P}_{\tau}p_{t}\left(\mathbf{x}\right)\nonumber \\
 & = & \int\mathrm{d}\mathbf{y}\ p\left(\mathbf{y},\mathbf{x};\tau\right)p_{t}\left(\mathbf{y}\right)
\end{eqnarray}
and
\begin{eqnarray}
u_{t+\tau}\left(\mathbf{x}\right) & = & \mathcal{T}_{\tau}u_{t}\left(\mathbf{x}\right)\nonumber \\
 & = & \int\mathrm{d}\mathbf{y}\ \frac{\mu\left(\mathbf{y}\right)}{\mu\left(\mathbf{x}\right)}p\left(\mathbf{y},\mathbf{x};\tau\right)u_{t}\left(\mathbf{y}\right),
\end{eqnarray}
where $p_{t}\left(\mathbf{x}\right)$ denotes the probability density
of $\mathbf{x}_{t}$ and $u_{t}\left(\mathbf{x}\right)=\mu\left(\mathbf{x}\right)^{-1}p_{t}\left(\mathbf{x}\right)$
denotes the density weighted by the inverse of the stationary density.
The relationship between the three operators can be summarized as
follows:
\begin{enumerate}
\item $\mathcal{K}_{\tau}$ is adjoint to $\mathcal{T}_{\tau}$ in the sense
of
\begin{equation}
\left\langle \mathcal{K}_{\tau}f_{1},f_{2}\right\rangle _{\mu}=\left\langle f_{1},\mathcal{T}_{\tau}f_{2}\right\rangle _{\mu}
\end{equation}
for any $f_{1},f_{2}\in L_{\mu}^{2}$. If $\{\mathbf{x}_{t}\}$ is
reversible, $\mathcal{K}_{\tau}$ and $\mathcal{T}_{\tau}$ are self-adjoint
with respect to $\left\langle \cdot,\cdot\right\rangle _{\mu}$, i.e.,
$\mathcal{K}_{\tau}=\mathcal{T}_{\tau}$.
\item Defining the multiplication operator $\mathcal{M}_{\mu}:L_{\mu}^{2}\mapsto L_{\mu^{-1}}^{2}$
as $\mathcal{M}_{\mu}f\left(\mathbf{x}\right)=\mu\left(\mathbf{x}\right)\cdot f\left(\mathbf{x}\right)$,
the Markov propagator $\mathcal{P}_{\tau}$ can be expressed as 
\begin{equation}
\mathcal{P}_{\tau}=\mathcal{M}_{\mu}\mathcal{T}_{\tau}\mathcal{M}_{\mu}^{-1}.
\end{equation}
Under the detailed balance condition, $\mathcal{P}_{\tau}$ is self-adjoint
with respect to $\left\langle \cdot,\cdot\right\rangle _{\mu^{-1}}$.
\end{enumerate}
We can also find the finite-dimensional approximation $\mathcal{P}_{\tau}\chi_{i}\approx\mathbf{p}_{i}^{\top}\boldsymbol{\chi}$
and $\mathcal{T}_{\tau}\chi_{i}\approx\mathbf{t}_{i}^{\top}\boldsymbol{\chi}$
of $\mathcal{P}_{\tau}$ and $\mathcal{T}_{\tau}$ by minimizing errors
$\sum_{i}\left\langle \mathbf{p}_{i}^{\top}\boldsymbol{\chi}-\mathcal{P}_{\tau}\chi_{i},\mathbf{p}_{i}^{\top}\boldsymbol{\chi}-\mathcal{P}_{\tau}\chi_{i}\right\rangle _{\omega}$
and $\sum_{i}\left\langle \mathbf{t}_{i}^{\top}\boldsymbol{\chi}-\mathcal{T}_{\tau}\chi_{i},\mathbf{t}_{i}^{\top}\boldsymbol{\chi}-\mathcal{T}_{\tau}\chi_{i}\right\rangle _{\omega}$
for some weight function $\omega\left(\mathbf{x}\right)$. However,
it is still unknown how to select the weight functions so that the
approximation errors can be easily computed from simulation data as
in the approximation of $\mathcal{K}_{\tau}$. For example, if we
select $\omega\left(\mathbf{x}\right)=\rho\left(\mathbf{x}\right)^{-1}$,
the approximation error of $\mathcal{P}_{\tau}$ is
\begin{eqnarray}
\sum_{i}\left\langle \mathbf{p}_{i}^{\top}\boldsymbol{\chi}-\mathcal{P}_{\tau}\chi_{i},\mathbf{p}_{i}^{\top}\boldsymbol{\chi}-\mathcal{P}_{\tau}\chi_{i}\right\rangle _{\rho^{-1}} & = & \sum_{i}\left\langle \mathbf{p}_{i}^{\top}\boldsymbol{\chi},\mathbf{p}_{i}^{\top}\boldsymbol{\chi}\right\rangle _{\rho^{-1}}-2\sum_{i}\left\langle \mathbf{p}_{i}^{\top}\boldsymbol{\chi},\mathcal{P}_{\tau}\chi_{i}\right\rangle _{\rho^{-1}}\nonumber \\
 &  & +\sum_{i}\left\langle \mathcal{P}_{\tau}\chi_{i},\mathcal{P}_{\tau}\chi_{i}\right\rangle _{\rho^{-1}}\nonumber \\
 & = & \sum_{i}\mathbb{E}_{\rho}\left[\frac{\mathbf{p}_{i}^{\top}\boldsymbol{\chi}\left(\mathbf{x}_{t}\right)\boldsymbol{\chi}\left(\mathbf{x}_{t}\right)^{\top}\mathbf{p}_{i}}{\rho\left(\mathbf{x}_{t}\right)^{2}}\right]\nonumber \\
 &  & -2\sum_{i}\mathbb{E}_{\rho}\left[\frac{\mathbf{p}_{i}^{\top}\boldsymbol{\chi}\left(\mathbf{x}_{t+\tau}\right)\chi_{i}\left(\mathbf{x}_{t}\right)}{\rho\left(\mathbf{x}_{t+\tau}\right)\rho\left(\mathbf{x}_{t}\right)}\right]\nonumber \\
 &  & +\sum_{i}\left\langle \mathcal{P}_{\tau}\chi_{i},\mathcal{P}_{\tau}\chi_{i}\right\rangle _{\rho^{-1}}
\end{eqnarray}
where the last term on the right hand side is a constant independent
of $\mathbf{p}_{i}$, and the computation of the first two terms is
infeasible for unknown $\rho$. For $\mathcal{T}_{\tau}$, the weight
function is generally set to be $\omega=\rho$, and the corresponding
approximation error is then
\begin{eqnarray}
\sum_{i}\left\langle \mathbf{t}_{i}^{\top}\boldsymbol{\chi}-\mathcal{T}_{\tau}\chi_{i},\mathbf{t}_{i}^{\top}\boldsymbol{\chi}-\mathcal{T}_{\tau}\chi_{i}\right\rangle _{\rho} & = & \sum_{i}\left\langle \mathbf{t}_{i}^{\top}\boldsymbol{\chi},\mathbf{t}_{i}^{\top}\boldsymbol{\chi}\right\rangle _{\rho}-2\sum_{i}\left\langle \mathbf{t}_{i}^{\top}\boldsymbol{\chi},\mathcal{T}_{\tau}\chi_{i}\right\rangle _{\rho}\nonumber \\
 &  & +\sum_{i}\left\langle \mathcal{T}_{\tau}\chi_{i},\mathcal{T}_{\tau}\chi_{i}\right\rangle _{\rho}\nonumber \\
 & = & \sum_{i}\mathbb{E}_{\rho}\left[\mathbf{t}_{i}^{\top}\boldsymbol{\chi}\left(\mathbf{x}_{t}\right)\boldsymbol{\chi}\left(\mathbf{x}_{t}\right)^{\top}\mathbf{t}_{i}\right]\nonumber \\
 &  & -2\sum_{i}\mathbb{E}_{\rho}\left[\frac{\rho\left(\mathbf{x}_{t+\tau}\right)\mu\left(\mathbf{x}_{t}\right)}{\mu\left(\mathbf{x}_{t+\tau}\right)\rho\left(\mathbf{x}_{t}\right)}\cdot\mathbf{t}_{i}^{\top}\boldsymbol{\chi}\left(\mathbf{x}_{t+\tau}\right)\chi_{i}\left(\mathbf{x}_{t}\right)\right]\nonumber \\
 &  & +\sum_{i}\left\langle \mathcal{T}_{\tau}\chi_{i},\mathcal{T}_{\tau}\chi_{i}\right\rangle _{\rho}
\end{eqnarray}
which is difficult to estimate unless the empirical distribution $\rho$
is consistent with $\mu$ or the system is reversible. (For reversible
systems, $\mathcal{K}_{\tau}=\mathcal{T}_{\tau}$ and the finite-dimensional
approximation of $\mathcal{K}_{\tau}$ is therefore also that of $\mathcal{T}_{\tau}$.)
In general cases, only the Koopman operator can be reliably estimated
from the non-equilibrium data.

\section{Properties of the empirical distribution\label{sec:Properties-of-rho}}

We first consider the case where the simulation data consist of $M$
independent trajectories $\{\mathbf{x}_{t}^{1}\}_{t=1}^{T},\ldots,\{\mathbf{x}_{t}^{K}\}_{t=1}^{T}$
of length $T$ and the initial state $x_{0}^{k}\stackrel{\mathrm{iid}}{\sim}p_{0}\left(\mathbf{x}\right)$.
In this case, $\rho$ can be given by
\begin{equation}
\rho=\frac{1}{T-\tau}\sum_{t=1}^{T-\tau}\mathcal{P}_{t}p_{0}
\end{equation}
where $\mathcal{P}_{t}$ denotes the forward operator defined in Appendix
\ref{sec:Dynamical-operators}.

For an arbitrary function $h$ of $\mathbf{x}_{t}$ and $\mathbf{x}_{t+\tau}$,
we have
\begin{eqnarray}
\mathbb{E}\left[\frac{1}{M\left(T-\tau\right)}\sum_{k=1}^{K}\sum_{t=1}^{T-\tau}h\left(\mathbf{x}_{t}^{k},\mathbf{x}_{t+\tau}^{k}\right)\right] & = & \frac{1}{T-\tau}\sum_{t=1}^{T-\tau}\mathbb{E}_{\mathcal{P}_{t}p_{0}}\left[h\left(\mathbf{x}_{t},\mathbf{x}_{t+\tau}\right)\right]\nonumber \\
 & = & \mathbb{E}_{\rho}\left[h\left(\mathbf{x}_{t},\mathbf{x}_{t+\tau}\right)\right]
\end{eqnarray}
and
\begin{eqnarray}
\frac{1}{M\left(T-\tau\right)}\sum_{k=1}^{K}\sum_{t=1}^{T-\tau}h\left(\mathbf{x}_{t}^{k},\mathbf{x}_{t+\tau}^{k}\right) & = & \frac{1}{T-\tau}\sum_{t=1}^{T-\tau}\left(\frac{1}{K}\sum_{k=1}^{K}h\left(\mathbf{x}_{t}^{k},\mathbf{x}_{t+\tau}^{k}\right)\right)\nonumber \\
 & \stackrel{p}{\to} & \frac{1}{T-\tau}\sum_{t=1}^{T-\tau}\mathbb{E}_{\mathcal{P}_{t}p_{0}}\left[h\left(\mathbf{x}_{t},\mathbf{x}_{t+\tau}\right)\right]\nonumber \\
 & = & \mathbb{E}_{\rho}\left[h\left(\mathbf{x}_{t},\mathbf{x}_{t+\tau}\right)\right]
\end{eqnarray}
as $M\to\infty$, where ``$\stackrel{p}{\to}$'' denotes the convergence
in probability. Therefore $\hat{\mathbf{C}}\left(0\right)$ and $\hat{\mathbf{C}}\left(\tau\right)$
are unbiased and consistent estimates of
\begin{eqnarray}
\mathbf{C}_{\rho}\left(0\right) & = & \mathbb{E}_{\rho}\left[\boldsymbol{\chi}\left(\mathbf{x}_{t}\right)\boldsymbol{\chi}\left(\mathbf{x}_{t}\right)^{\top}\right]\\
\mathbf{C}_{\rho}\left(\tau\right) & = & \mathbb{E}_{\rho}\left[\boldsymbol{\chi}\left(\mathbf{x}_{t}\right)\boldsymbol{\chi}\left(\mathbf{x}_{t+\tau}\right)^{\top}\right]
\end{eqnarray}
The importance sampling approximation provided by (\ref{eq:reweighting})
is also consistent because
\begin{eqnarray}
\frac{\sum_{k=1}^{K}\sum_{t=1}^{T-\tau}w\left(\mathbf{x}_{t}^{k}\right)h\left(\mathbf{x}_{t}^{k},\mathbf{x}_{t+\tau}^{k}\right)}{\sum_{k=1}^{K}\sum_{t=1}^{T-\tau}w\left(\mathbf{x}_{t}^{k}\right)} & = & \frac{\frac{1}{K\left(T-\tau\right)}\sum_{k=1}^{K}\sum_{t=1}^{T-\tau}w\left(\mathbf{x}_{t}^{k}\right)h\left(\mathbf{x}_{t}^{k},\mathbf{x}_{t+\tau}^{k}\right)}{\frac{1}{K\left(T-\tau\right)}\sum_{k=1}^{K}\sum_{t=1}^{T-\tau}w\left(\mathbf{x}_{t}^{k}\right)}\nonumber \\
 & \stackrel{p}{\to} & \frac{\mathbb{E}_{\rho}\left[w\left(\mathbf{x}_{t}\right)h\left(\mathbf{x}_{t},\mathbf{x}_{t+\tau}\right)\right]}{\mathbb{E}_{\rho}\left[w\left(\mathbf{x}_{t}\right)\right]}\nonumber \\
 & = & \frac{\iint\mathrm{d}\mathbf{x}\mathrm{d}\mathbf{y}\ \frac{\mu\left(\mathbf{x}\right)}{\rho\left(\mathbf{x}\right)}\rho\left(\mathbf{x}\right)p\left(\mathbf{x},\mathbf{y}\right)h\left(\mathbf{x}_{t},\mathbf{x}_{t+\tau}\right)}{\int\mathrm{d}\mathbf{x}\ \frac{\mu\left(\mathbf{x}\right)}{\rho\left(\mathbf{x}\right)}\rho\left(\mathbf{x}\right)}\nonumber \\
 & = & \mathbb{E}_{\mu}\left[h\left(\mathbf{x}_{t},\mathbf{x}_{t+\tau}\right)\right]
\end{eqnarray}
If we further assume that the finite-dimensional approximation (\ref{eq:eigenfunction-approximation})
of spectral components is exact, i.e., $\{\mathbf{x}_{t}\}$ has only
$m$ nonzero eigenvalues and $\psi_{i}=\mathbf{b}_{i}^{\top}\boldsymbol{\chi}$
holds exactly for $i=1,\ldots,m$, we can get
\begin{eqnarray}
\frac{\rho\left(\mathbf{x}\right)}{\mu\left(\mathbf{x}\right)} & = & \frac{1}{T-\tau}\sum_{t=1}^{T-\tau}\frac{\mathcal{P}_{t}p_{0}\left(\mathbf{x}\right)}{\mu\left(\mathbf{x}\right)}\nonumber \\
 & = & \frac{1}{T-\tau}\sum_{t=1}^{T-\tau}\frac{\sum_{i=1}^{m}\lambda_{i}\left(t\right)\,\mu\left(\mathbf{x}\right)\psi_{i}\left(\mathbf{x}\right)\,\left\langle \psi_{i},p_{0}\right\rangle }{\mu\left(\mathbf{x}\right)}\nonumber \\
 & = & \left[\frac{\sum_{t=1}^{T-\tau}\lambda_{1}\left(t\right)}{T-\tau}\left\langle \psi_{1},p_{0}\right\rangle ,\ldots,\frac{\sum_{t=1}^{T-\tau}\lambda_{m}\left(t\right)}{T-\tau}\left\langle \psi_{m},p_{0}\right\rangle \right]\mathbf{B}^{\top}\boldsymbol{\chi}\left(\mathbf{x}\right)
\end{eqnarray}
which implies that (\ref{eq:Koopman-reweighting}) can be exactly
satisfied with
\begin{equation}
\mathbf{u}=\mathbf{B}\left[\frac{\sum_{t=1}^{T-\tau}\lambda_{1}\left(t\right)}{T-\tau}\left\langle \psi_{1},p_{0}\right\rangle ,\ldots,\frac{\sum_{t=1}^{T-\tau}\lambda_{m}\left(t\right)}{T-\tau}\left\langle \psi_{m},p_{0}\right\rangle \right]^{\top}
\end{equation}
Moreover, under the finite-dimensional assumption, we have
\begin{eqnarray}
\mathcal{K}_{\tau}\chi_{i}\left(\mathbf{x}\right) & = & \int\mathrm{d}\mathbf{y}\ p\left(\mathbf{x},\mathbf{y};\tau\right)\chi_{i}\left(\mathbf{y}\right)\nonumber \\
 & = & \int\mathrm{d}\mathbf{y}\ \sum_{i=1}^{m}\lambda_{i}\left(t\right)\mu\left(\mathbf{y}\right)\psi_{i}\left(\mathbf{y}\right)\psi_{i}\left(\mathbf{x}\right)\chi_{i}\left(\mathbf{y}\right)\nonumber \\
 & = & \left(\int\mathrm{d}\mathbf{y}\ \sum_{i=1}^{m}\lambda_{i}\left(t\right)\mu\left(\mathbf{y}\right)\psi_{i}\left(\mathbf{y}\right)\chi_{i}\left(\mathbf{y}\right)\right)\mathbf{b}_{i}^{\top}\boldsymbol{\chi}\left(\mathbf{x}\right)
\end{eqnarray}
Thus there exists a matrix $\mathbf{K}=(\mathbf{k}_{1},\ldots,\mathbf{k}_{m})$
so that $\mathcal{K}_{\tau}\boldsymbol{\chi}=\mathbf{K}^{\top}\boldsymbol{\chi}$
holds exactly with $\mathcal{K}_{\tau}\boldsymbol{\chi}=\left(\mathcal{K}_{\tau}\chi_{1},\ldots,\mathcal{K}_{\tau}\chi_{m}\right)^{\top}$.
Considering that
\begin{eqnarray}
\mathbb{E}_{\mu}\left[\boldsymbol{\chi}\left(\mathbf{x}_{t+\tau}\right)\right] & = & \mathbb{E}_{\mu}\left[\mathcal{K}_{\tau}\boldsymbol{\chi}\left(\mathbf{x}_{t}\right)\right]\nonumber \\
 & = & \mathbb{E}_{\mu}\left[\mathbf{K}^{\top}\boldsymbol{\chi}\left(\mathbf{x}_{t}\right)\right]\nonumber \\
 & = & \int\mathrm{d}\mathbf{x}\ \mathbf{u}^{\top}\boldsymbol{\chi}\left(\mathbf{x}\right)\cdot\rho\left(\mathbf{x}\right)\cdot\mathbf{K}^{\top}\boldsymbol{\chi}\left(\mathbf{x}\right)\nonumber \\
 & = & \mathbf{K}^{\top}\left(\int\mathrm{d}\mathbf{x}\ \rho\left(\mathbf{x}\right)\boldsymbol{\chi}\left(\mathbf{x}\right)\boldsymbol{\chi}\left(\mathbf{x}\right)^{\top}\right)\mathbf{u}\nonumber \\
 & = & \mathbf{K}^{\top}\mathbf{C}_{\rho}\left(0\right)\mathbf{u}
\end{eqnarray}
and

\begin{eqnarray}
\mathbb{E}_{\mu}\left[\boldsymbol{\chi}\left(\mathbf{x}_{t}\right)\right] & = & \int\mathrm{d}\mathbf{x}\ \mathbf{u}^{\top}\boldsymbol{\chi}\left(\mathbf{x}\right)\cdot\rho\left(\mathbf{x}\right)\cdot\boldsymbol{\chi}\left(\mathbf{x}\right)\nonumber \\
 & = & \mathbf{C}_{\rho}\left(0\right)\mathbf{u},
\end{eqnarray}
we can obtain from $\mathbb{E}_{\mu}\left[\boldsymbol{\chi}\left(\mathbf{x}_{t+\tau}\right)\right]=\mathbb{E}_{\mu}\left[\boldsymbol{\chi}\left(\mathbf{x}_{t}\right)\right]$
that
\begin{equation}
\mathbf{C}_{\rho}\left(0\right)^{-1}\mathbf{K}^{\top}\mathbf{C}_{\rho}\left(0\right)\mathbf{u}=\mathbf{u}.
\end{equation}
Since $\mathbf{C}_{\rho}\left(0\right)^{-1}\mathbf{K}^{\top}\mathbf{C}_{\rho}\left(0\right)$
is similar to $\mathbf{K}^{\top}$ and the largest eigenvalue of $\mathbf{K}$
is $1$, we can conclude that $\mathbf{u}$ is the eigenvector of
$\mathbf{C}_{\rho}\left(0\right)^{-1}\mathbf{K}^{\top}\mathbf{C}_{\rho}\left(0\right)$
with the largest eigenvalue.

In more general cases, where, for example, trajectories are generated
with different initial conditions and different lengths, the similar
conclusions can be obtained by considering that
\begin{equation}
\frac{\mathcal{P}_{t}p_{0}\left(\mathbf{x}\right)}{\mu\left(\mathbf{x}\right)}=\left[\lambda_{1}\left(t\right)\left\langle \psi_{1},p_{0}\right\rangle ,\ldots,\lambda_{m}\left(t\right)\left\langle \psi_{m},p_{0}\right\rangle \right]\mathbf{B}^{\top}\boldsymbol{\chi}\left(\mathbf{x}\right)\in\mathrm{span}\{\chi_{1},\ldots,\chi_{m}\}
\end{equation}
for all $p_{0}$ and $t$ if the finite-dimensional approximation
(\ref{eq:eigenfunction-approximation}) is assumed to be exact, i.e.,
the ratio between $\rho$ and $\mu$ can always be expressed as a
linear combination of $\boldsymbol{\chi}$ under this assumption.

\section{Limit of the Koopman model approximation error\label{sec:Limit-of-the-error}}

The mean square error of the nonreversible Koopman model approximation
is (see \cite{williams2015data})
\begin{equation}
\mathrm{MSE}=\frac{1}{N}\sum_{t=1}^{T-\tau}\sum_{i=1}^{m}\left\Vert \mathbf{k}_{i}^{\top}\chi_{i}\left(\mathbf{x}_{t}\right)-\chi_{i}\left(\mathbf{x}_{t+\tau}\right)\right\Vert ^{2}
\end{equation}
Under the condition $N\to\infty$, we have
\begin{eqnarray*}
\mathrm{MSE} & = & \sum_{i=1}^{m}\int\mathrm{d}\mathbf{x}\ \rho\left(\mathbf{x}\right)\left(\mathbf{k}_{i}^{\top}\boldsymbol{\chi}-\mathcal{K}_{\tau}\chi_{i}\right)^{\top}\left(\mathbf{k}_{i}^{\top}\boldsymbol{\chi}-\mathcal{K}_{\tau}\chi_{i}\right)\\
 & = & \sum_{i=1}^{m}\left\langle \mathbf{k}_{i}^{\top}\boldsymbol{\chi}-\mathcal{K}_{\tau}\chi_{i},\mathbf{k}_{i}^{\top}\boldsymbol{\chi}-\mathcal{K}_{\tau}\chi_{i}\right\rangle _{\rho}
\end{eqnarray*}

\section{Proof of (\ref{eq:Sigma}) and (\ref{eq:edmd-symmetrization})}

\label{sec:Proof-of-symmetrization}If $\{\mathbf{x}_{t}\}$ is a
time-reversible stochastic process, the time-lagged cross correlation
between two arbitrary observable quantities $f_{1}\left(\mathbf{x}_{t}\right)$
and $f_{2}\left(\mathbf{x}_{t}\right)$ at equilibrium is symmetric
in the sense of $\mathbb{E}_{\mu}\left[f_{1}\left(\mathbf{x}_{t}\right)f_{2}\left(\mathbf{x}_{t+\tau}\right)\right]=\mathbb{E}_{\mu}\left[f_{2}\left(\mathbf{x}_{t}\right)f_{1}\left(\mathbf{x}_{t+\tau}\right)\right]$.
We can obtain that
\begin{eqnarray*}
\mathbf{C}\left(0\right) & = & \mathbb{E}_{\mu}\left[\boldsymbol{\chi}\left(x_{t}\right)\boldsymbol{\chi}\left(x_{t}\right)^{\top}\right]\\
 & = & \frac{1}{2}\mathbb{E}_{\mu}\left[\boldsymbol{\chi}\left(x_{t}\right)\boldsymbol{\chi}\left(x_{t}\right)^{\top}+\boldsymbol{\chi}\left(x_{t+\tau}\right)\boldsymbol{\chi}\left(x_{t+\tau}\right)^{\top}\right]\\
 & \approx & \frac{1}{2}\sum_{t=1}^{T-\tau}w\left(\mathbf{x}_{t}\right)\left(\boldsymbol{\chi}\left(x_{t}\right)\boldsymbol{\chi}\left(x_{t}\right)^{\top}+\boldsymbol{\chi}\left(x_{t+\tau}\right)\boldsymbol{\chi}\left(x_{t+\tau}\right)^{\top}\right)
\end{eqnarray*}
and
\begin{eqnarray*}
\mathbf{C}\left(\tau\right) & = & \mathbb{E}_{\mu}\left[\boldsymbol{\chi}\left(x_{t}\right)\boldsymbol{\chi}\left(x_{t+\tau}\right)^{\top}\right]\\
 & = & \frac{1}{2}\mathbb{E}_{\mu}\left[\boldsymbol{\chi}\left(x_{t}\right)\boldsymbol{\chi}\left(x_{t+\tau}\right)^{\top}+\boldsymbol{\chi}\left(x_{t+\tau}\right)\boldsymbol{\chi}\left(x_{t}\right)^{\top}\right]\\
 & \approx & \frac{1}{2}\sum_{t=1}^{T-\tau}w\left(\mathbf{x}_{t}\right)\left(\boldsymbol{\chi}\left(x_{t}\right)\boldsymbol{\chi}\left(x_{t+\tau}\right)^{\top}+\boldsymbol{\chi}\left(x_{t+\tau}\right)\boldsymbol{\chi}\left(x_{t}\right)^{\top}\right)
\end{eqnarray*}
where the approximation steps in the above equations come from (\ref{eq:reweighting}).

\section{Analysis of the reversible estimator\label{sec:Analysis-of-reversibility}}

Here we analyze properties of the reversible estimator under the following
assumptions:
\begin{assumption}
\label{assu:constant-function}The constant function $\mathbbm1\in\mathrm{span}\{\chi_{1},\ldots,\chi_{m}\}$,
i.e., there is a vector $\mathbf{v}$ so that $\mathbf{v}^{\top}\boldsymbol{\chi}=\mathbbm1$.
\end{assumption}

\begin{assumption}
\label{assu:positive-definite}$\hat{\mathbf{C}}\left(0\right)$,
$\hat{\mathbf{C}}_{\mathrm{rev}}\left(0\right)$ are positive-definite,
and all weights $w\left(\mathbf{x}_{t}\right)$ are positive.
\end{assumption}
Under Assumption \ref{assu:positive-definite}, $\mathbf{K}_{\mathrm{rev}}$
is similar to
\begin{equation}
\hat{\mathbf{C}}_{\mathrm{rev}}\left(0\right)^{\frac{1}{2}}\mathbf{K}_{\mathrm{rev}}\hat{\mathbf{C}}_{\mathrm{rev}}\left(0\right)^{-\frac{1}{2}}=\hat{\mathbf{C}}_{\mathrm{rev}}\left(0\right)^{-\frac{1}{2}}\hat{\mathbf{C}}_{\mathrm{rev}}\left(\tau\right)\hat{\mathbf{C}}_{\mathrm{rev}}\left(0\right)^{-\frac{1}{2}}
\end{equation}
where $\hat{\mathbf{C}}_{\mathrm{rev}}\left(0\right)^{\frac{1}{2}}$
denotes the symmetric square root of $\hat{\mathbf{C}}_{\mathrm{rev}}\left(0\right)$.
Therefore the eigenvalue problem of $\mathbf{K}_{\mathrm{rev}}$ can
be solved in the real field. In addition, for any $\lambda$ and nonzero
$\mathbf{b}$ which satisfy $\mathbf{K}_{\mathrm{rev}}\mathbf{b}=\lambda\mathbf{b}$,
we have
\begin{eqnarray*}
\left|\lambda\right| & = & \left|\frac{\mathbf{b}^{\top}\hat{\mathbf{C}}_{\mathrm{rev}}\left(\tau\right)\mathbf{b}}{\mathbf{b}^{\top}\hat{\mathbf{C}}_{\mathrm{rev}}\left(0\right)\mathbf{b}}\right|\\
 & = & \frac{\left|\mathbf{b}^{\top}\mathbf{X}^{\top}\mathbf{W}\mathbf{Y}\mathbf{b}+\mathbf{b}^{\top}\mathbf{Y}^{\top}\mathbf{W}\mathbf{X}\mathbf{b}\right|}{\mathbf{b}^{\top}\mathbf{X}^{\top}\mathbf{W}\mathbf{X}\mathbf{b}+\mathbf{b}^{\top}\mathbf{Y}^{\top}\mathbf{W}\mathbf{Y}\mathbf{b}}\\
 & \le & 1
\end{eqnarray*}
which implies that the spectral radius of $\mathbf{K}_{\mathrm{rev}}$
is not larger than $1$.

Under Assumption \ref{assu:constant-function}, the matrix $\mathbf{K}$
given by (\ref{eq:K-approximation}) satisfies
\begin{eqnarray}
\mathbf{K}\mathbf{v} & = & \hat{\mathbf{C}}\left(0\right)^{-1}\hat{\mathbf{C}}\left(\tau\right)\mathbf{v}\nonumber \\
 & = & \hat{\mathbf{C}}\left(0\right)^{-1}\left(\frac{1}{N}\mathbf{X}^{\top}\mathbf{Y}\mathbf{v}\right)\nonumber \\
 & = & \hat{\mathbf{C}}\left(0\right)^{-1}\left(\frac{1}{N}\mathbf{X}^{\top}\mathbf{X}\mathbf{v}\right)\nonumber \\
 & = & \hat{\mathbf{C}}\left(0\right)^{-1}\hat{\mathbf{C}}\left(0\right)\mathbf{v}\nonumber \\
 & = & \mathbf{v}
\end{eqnarray}
So $1$ is an eigenvalue of $\mathbf{K}$ and the eigenvalue problem
(\ref{eq:u-direction}) can be exactly solved.

We now show that the weight function $w\left(\mathbf{x}\right)$ remains
fixed after replacing $\mathbf{K}$ by $\mathbf{K}_{\mathrm{rev}}$,
i.e.,
\begin{equation}
\hat{\mathbf{C}}\left(0\right)^{-1}\mathbf{K}_{\mathrm{rev}}^{\top}\hat{\mathbf{C}}\left(0\right)\mathbf{u}=\mathbf{u}
\end{equation}
for $\mathbf{u}$ satisfying $\mathbf{1}^{\top}\mathbf{X}\mathbf{u}=1$
and
\begin{equation}
\hat{\mathbf{C}}\left(0\right)^{-1}\mathbf{K}^{\top}\hat{\mathbf{C}}\left(0\right)\mathbf{u}=\mathbf{u}\Leftrightarrow\mathbf{Y}^{\top}\mathbf{X}\mathbf{u}=\mathbf{X}^{\top}\mathbf{X}\mathbf{u}
\end{equation}
Considering that
\begin{eqnarray}
\hat{\mathbf{C}}_{\mathrm{rev}}\left(0\right)\mathbf{v} & = & \frac{1}{2}\left(\mathbf{X}^{\top}\mathbf{W}\mathbf{X}+\mathbf{Y}^{\top}\mathbf{W}\mathbf{Y}\right)\mathbf{v}\nonumber \\
 & = & \frac{1}{2}\left(\mathbf{X}^{\top}\mathbf{X}\mathbf{u}+\mathbf{Y}^{\top}\mathbf{X}\mathbf{u}\right)\nonumber \\
 & = & N\hat{\mathbf{C}}\left(0\right)\mathbf{u}
\end{eqnarray}
Thus,
\begin{eqnarray}
\hat{\mathbf{C}}\left(0\right)^{-1}\mathbf{K}_{\mathrm{rev}}^{\top}\hat{\mathbf{C}}\left(0\right)\mathbf{u} & = & \frac{1}{N}\hat{\mathbf{C}}\left(0\right)^{-1}\hat{\mathbf{C}}_{\mathrm{rev}}\left(\tau\right)\hat{\mathbf{C}}_{\mathrm{rev}}\left(0\right)^{-1}\hat{\mathbf{C}}_{\mathrm{rev}}\left(0\right)\mathbf{v}\nonumber \\
 & = & \frac{1}{2N}\hat{\mathbf{C}}\left(0\right)^{-1}\left(\mathbf{X}^{\top}\mathbf{W}\mathbf{Y}+\mathbf{Y}^{\top}\mathbf{W}\mathbf{X}\right)\mathbf{v}\nonumber \\
 & = & \mathbf{u}
\end{eqnarray}

\section{De-correlation of basis functions\label{sec:Detailed-decorreltion-procedure}}

In Section \ref{subsec:edmd} and Algorithm 1, the basis functions
$\boldsymbol{\chi}$ are assumed to be linearly independent on the
sampled data so that $\hat{\mathbf{C}}\left(0\right)$ is invertible
and the matrix $\mathbf{K}$ given in (\ref{eq:K-approximation})
is well defined. In some publications, e.g. \cite{williams2015data},
$\mathbf{K}$ is calculated as $\mathbf{K}=\hat{\mathbf{C}}\left(0\right)^{\dagger}\hat{\mathbf{C}}\left(\tau\right)$
by using the pseudo-inverse $\hat{\mathbf{C}}\left(0\right)^{\dagger}$
of $\hat{\mathbf{C}}\left(0\right)$, however this approach cannot
completely avoid numerical instabilities. In this paper, we utilize
principal component analysis (PCA) \cite{Pearson_PhilMag1901_PCA}
to explicitly reduce correlations between basis functions as follows:
First, we compute the empirical mean of basis functions and the covariance
matrix of mean-centered basis functions:
\begin{eqnarray}
\boldsymbol{\pi} & = & \frac{1}{N}\mathbf{X}^{\top}\mathbf{1}\label{eq:pi}\\
\mathrm{COV} & = & \frac{1}{N}\mathbf{X}^{\top}\mathbf{X}-\boldsymbol{\pi}\boldsymbol{\pi}^{\top}\label{eq:cov}
\end{eqnarray}
Next, perform the truncated eigendecomposition of the covariance matrix
as
\begin{equation}
\mathrm{COV}\approx\mathbf{Q}_{d}^{\top}\mathbf{S}_{d}\mathbf{Q}_{d},
\end{equation}
where the diagonal of matrix $\mathbf{S}_{d}$ contains all positive
eigenvalues that are larger than $\epsilon_{0}$ and absolute values
of all negative eigenvalues ($\epsilon_{0}=10^{-10}$ in our applications).
Last, the new basis functions are given by
\begin{equation}
\boldsymbol{\chi}^{\mathrm{new}}=\left[\begin{array}{c}
\mathbf{Q}_{d}^{\top}\mathbf{S}_{d}^{\frac{1}{2}}\left(\boldsymbol{\chi}-\boldsymbol{\pi}\right)\\
\mathbbm1
\end{array}\right]\label{eq:decorrelation}
\end{equation}
Here $\mathbf{Q}_{d}^{\top}\mathbf{S}_{d}^{\frac{1}{2}}\left(\boldsymbol{\chi}-\boldsymbol{\pi}\right)$
is the PCA whitening transformation of the original basis functions,
which transforms $\boldsymbol{\chi}$ into all available principal
components and scales each component to have a variance of $1$, and
the constant function $\mathbbm1$ is added to the basis function
so that the eigenvalue problem (\ref{eq:u-direction}) in the estimation
of equilibrium distribution can be exactly solved (see Appendix \ref{sec:Analysis-of-reversibility}).
It can be verified that the direct estimate $\hat{\mathbf{C}}\left(0\right)$
of the covariance matrix obtained from $\boldsymbol{\chi}^{\mathrm{new}}\left(\mathbf{x}_{t}\right)$
is an identity matrix. The corresponding estimate of the Koopman operator
is given by $\mathbf{K}=\hat{\mathbf{C}}\left(0\right)=\frac{1}{N}\mathbf{X}^{\top}\mathbf{Y}$.

For convenience of notation, we denote the transformation (\ref{eq:decorrelation})
by
\begin{equation}
\boldsymbol{\chi}^{\mathrm{new}}=\mathrm{DC}\left[\boldsymbol{\chi}|\boldsymbol{\pi},\mathrm{COV}\right]
\end{equation}
Then the nonreversible Koopman estimation, which also perform EDMD,
linear VA and TICA in feature space, can be robustly implemented as
follows:

\noindent\textbf{Algorithm 1{*}: Nonreversible Koopman estimation
(with de-correlation of basis functions) }
\begin{enumerate}
\item Basis-transform input coordinates according to (\ref{eq:basis_transform}).
\item Compute $\boldsymbol{\pi}$ and $\mathrm{COV}$ by (\ref{eq:pi})
and (\ref{eq:cov}).
\item Let $\boldsymbol{\chi}:=\mathrm{DC}\left[\boldsymbol{\chi}|\boldsymbol{\pi},\mathrm{COV}\right]$,
and recalculate $\mathbf{X}$ and $\mathbf{Y}$ according to the new
basis functions.
\item Compute the matrix $\mathbf{K}=\frac{1}{N}\mathbf{X}^{\top}\mathbf{Y}$
and solve the eigenvalue problem $\mathbf{K}\mathbf{B}=\mathbf{B}\boldsymbol{\Lambda}$.
\item Output spectral components: Eigenvalues $\lambda_{i}$ and eigenfunctions
$\psi_{i}=\mathbf{b}_{i}^{\top}\boldsymbol{\chi}$.
\end{enumerate}
Furthermore, Koopman reweighting (Algorithm 2), can be robustly implemented
by the following algorithm:

\noindent\textbf{Algorithm 2{*}: Koopman reweighting (with de-correlation
of basis functions) }
\begin{enumerate}
\item Basis-transform input coordinates according to (\ref{eq:basis_transform}).
\item Compute $\mathbf{K}$ as in Algorithm 1{*}.
\item Compute $\mathbf{u}$ by solving $\mathbf{K}^{\top}\mathbf{u}=\mathbf{u}$
and normalize it by $\mathbf{1}^{\top}\mathbf{X}\mathbf{u}$.
\item Output weights: $w\left(\mathbf{x}_{t}\right)=\mathbf{x}_{t}^{\top}\mathbf{u}$.
\end{enumerate}
Similarly, we can also guarantee the positive-definiteness of $\hat{\mathbf{C}}_{\mathrm{rev}}\left(0\right)$
by de-correlation of basis functions based on the transformation
\begin{equation}
\boldsymbol{\chi}^{\mathrm{new}}=\mathrm{DC}\left[\boldsymbol{\chi}|\boldsymbol{\pi}_{\mathrm{eq}},\mathrm{COV}_{\mathrm{eq}}\right]
\end{equation}
where
\begin{eqnarray}
\boldsymbol{\pi}_{\mathrm{eq}} & = & \frac{1}{2}\left(\mathbf{X}+\mathbf{Y}\right)^{\top}\mathbf{W}\mathbf{1}\label{eq:pi-eq}\\
\mathrm{COV}_{\mathrm{eq}} & = & \frac{1}{2}\left(\mathbf{X}^{\top}\mathbf{W}\mathbf{X}+\mathbf{Y}^{\top}\mathbf{W}\mathbf{Y}\right)-\boldsymbol{\pi}_{\mathrm{eq}}\boldsymbol{\pi}_{\mathrm{eq}}^{\top}\label{eq:cov-eq}
\end{eqnarray}
are estimated equilibrium mean and covariance matrix of $\boldsymbol{\chi}$.
The corresponding reversible Koopman estimator which is consistent
with the variational approach, can be robustly implemented as follows:

\noindent\textbf{Algorithm 3{*}: Variational Koopman model and Eigendecomposition
(with de-correlation of basis functions) }
\begin{enumerate}
\item Basis-transform input coordinates according to (\ref{eq:basis_transform}).
\item Compute $\mathbf{u}$ as in Algorithm 2{*} and let $\mathbf{W}=\mathrm{diag}\left(\mathbf{X}\mathbf{u}\right)$.
\item Compute $\boldsymbol{\pi}_{\mathrm{eq}}$ and $\mathrm{COV}_{\mathrm{eq}}$
by (\ref{eq:pi-eq}) and (\ref{eq:cov-eq}).
\item Let $\boldsymbol{\chi}:=\mathrm{DC}\left[\boldsymbol{\chi}|\boldsymbol{\pi}_{\mathrm{eq}},\mathrm{COV}_{\mathrm{eq}}\right]$,
and recalculate $\mathbf{X}$ and $\mathbf{Y}$ according to the new
basis functions.
\item Compute $\mathbf{K}_{\mathrm{rev}}=\hat{\mathbf{C}}_{\mathrm{rev}}\left(\tau\right)=\frac{1}{2}\left(\mathbf{X}^{\top}\mathbf{W}\mathbf{Y}+\mathbf{Y}^{\top}\mathbf{W}\mathbf{X}\right)$
and solve the eigenvalue problem $\mathbf{K}_{\mathrm{rev}}\mathbf{B}=\mathbf{B}\boldsymbol{\Lambda}$.
\item Output spectral components: Eigenvalues $\lambda_{i}$ and eigenfunctions
$\psi_{i}=\mathbf{b}_{i}^{\top}\boldsymbol{\chi}$.
\end{enumerate}

\section{Simulation models and setups}

\subsection{One-dimensional diffusion process\label{subsec:One-dimensional-setup}}

The diffusion processes in Section \ref{subsec:One-dimensional-diffusion-process}
is driven by the Brownian dynamics
\begin{equation}
\mathrm{d}x_{t}=-\nabla U(x_{t})\mathrm{d}t+\sqrt{2\beta^{-1}}\mathrm{d}W_{t}\label{eq:brownian-dynamics-1d}
\end{equation}
where $\beta=0.3$, the time step is $0.002$, $x_{0}$ is uniformly
drawn in $[0,0.2]$, and the potential function is given by
\begin{equation}
U\left(x\right)=\frac{\sum_{i=1}^{5}\left(\left|x-c_{i}\right|+0.001\right)^{-2}u_{i}}{\sum_{i=1}^{5}\left(\left|x-c_{i}\right|+0.001\right)^{-2}}
\end{equation}
with $c_{1:5}=(-0.3,0.5,1,1.5,2.3)$. Simulations are implemented
by a reversibility preserving numerical discretization scheme proposed
in \cite{latorre2011structure} with bin size $0.02$. The basis functions
for estimators are chosen to be
\begin{equation}
\chi_{i}\left(x\right)=\exp\left(-\left(w_{i}x+b_{i}\right)^{2}\right),
\end{equation}
where $w_{i}$ and $b_{i}$ are randomly drawn in $[-1,1]$ and $[0,1]$.

\subsection{Two-dimensional diffusion process\label{subsec:Two-dimensional-setup}}

The dynamics of the two-dimensional diffusion process in Section \ref{subsec:Two-dimensional-diffusion-process}
has the same form as (\ref{eq:brownian-dynamics-1d}), where $\beta=0.5$,
sample interval is $0.05$, $\mathbf{x}_{0}=(x_{0},y_{0})$ is uniformly
drawn in $[-2,-1.5]\times[-1.5,2.5]$, and the potential function
is chosen as in \cite{MetznerSchuetteVandenEijnden_JCP06_TPT} by

\begin{eqnarray}
U\left(x,y\right) & = & 3\exp\left(-x^{2}-\left(y-\frac{1}{3}\right)^{2}\right)\nonumber \\
 &  & -3\exp\left(-x^{2}-\left(y-\frac{5}{3}\right)^{2}\right)\nonumber \\
 &  & -5\exp\left(-\left(x-1\right)^{2}-y^{2}\right)\nonumber \\
 &  & -5\exp\left(-\left(x+1\right)^{2}-y^{2}\right)\nonumber \\
 &  & +\frac{1}{5}x^{4}+\frac{1}{5}\left(y-\frac{1}{3}\right)^{4}.
\end{eqnarray}
Simulations are implemented by the same algorithm as in Appendix \ref{subsec:One-dimensional-setup}
with bin size $0.2\times0.2$. The basis functions for estimators
are also Gaussian functions
\begin{equation}
\chi_{i}\left(\mathbf{x}\right)=\exp\left(-\left(\mathbf{w}_{i}^{\top}\mathbf{x}+b_{i}\right)^{2}\right),
\end{equation}
with random weights $\mathbf{w}_{i}\in[-1,1]\times[-1,1]$ and $b_{i}\in[0,1]$.

\end{document}